\documentclass{article}

\usepackage{microtype}
\usepackage{graphicx}
\usepackage{booktabs} 
\usepackage{amsmath}
\usepackage{subfig}
\usepackage{caption} 
\usepackage{hyperref}

\newtheorem{theorem}{Theorem}[section]

\newtheorem{lemma}[theorem]{Lemma}

\usepackage{xcolor}
\numberwithin{equation}{section}
\newcommand{\R}{\mathbb{R}}
\usepackage{amssymb}
\usepackage{thm-restate}
\newcommand{\specialcell}[2][c]{%
  \begin{tabular}[#1]{@{}c@{}}#2\end{tabular}}

\usepackage[accepted]{icml2021}

\icmltitlerunning{Bayesian Optimization over Hybrid Spaces}

\begin{document}

\twocolumn[
\icmltitle{Bayesian Optimization over Hybrid Spaces}




\begin{icmlauthorlist}
\icmlauthor{Aryan Deshwal}{to}
\icmlauthor{Syrine Belakaria}{to}
\icmlauthor{Janardhan Rao Doppa}{to}
\end{icmlauthorlist}

\icmlaffiliation{to}{School of EECS, Washington State University, Pullman, USA}
\icmlcorrespondingauthor{Aryan Deshwal}{aryan.deshwal@wsu.edu}

\icmlkeywords{Machine Learning, ICML}

\vskip 0.3in
]



\printAffiliationsAndNotice{}  

\begin{abstract}
We consider the problem of optimizing hybrid structures (mixture of discrete and continuous input variables) via expensive black-box function evaluations. This problem arises in many real-world applications. For example, in materials design optimization via lab experiments, discrete and continuous variables correspond to the presence/absence of primitive elements and their relative concentrations respectively. The key challenge is to accurately model the complex interactions between discrete and continuous variables. In this paper, we propose a novel approach referred as {\em {\bf Hy}brid {\bf B}ayesian {\bf O}ptimization (HyBO)} by utilizing diffusion kernels, which are naturally defined over continuous and discrete variables. We develop a principled approach for constructing diffusion kernels over hybrid spaces by utilizing the additive kernel formulation, which allows additive interactions of all orders in a tractable manner. We theoretically analyze the modeling strength of additive hybrid kernels and prove that it has the {\em universal approximation} property. Our experiments on synthetic and six diverse real-world benchmarks show that HyBO significantly outperforms the state-of-the-art methods.
\end{abstract}

\vspace{-4.0ex}

\section{Introduction}

A large number of science and engineering applications involve optimizing hybrid spaces (mixture of discrete and continuous input variables) guided by expensive black-box function evaluations. For example, in materials design optimization, discrete variables correspond to the presence/absence of primitive elements and continuous variables correspond to their relative concentrations, and evaluation of each design involves performing an expensive physical lab experiment. A popular and effective framework for optimizing expensive black-box functions is Bayesian optimization (BO) \cite{BO-Survey:2016,bo_tutorial,greenhill2020bayesian,MESMO,ACDesign,USEMO,USEMOC,MESMOC,belakaria2020PSD}.  The key idea behind BO is to learn a surrogate statistical model and intelligently select the sequence of inputs for evaluation to approximately optimize the unknown objective. Gaussian process (GP) \cite{GP-Book} is the most popular choice for learning statistical models. GPs allow to incorporate domain knowledge about the problem in the form of a kernel over the input space and provide good uncertainty quantification. GPs have been successfully applied for both 
continuous \cite{BO-Survey:2016,MF-MESMO,iMOCA} and discrete spaces \cite{combo,MerCBO,reviewer_ref_3}. However, as we discuss in the related work section, there is very limited work on BO methods to optimize hybrid spaces \cite{SMAC,SMAC:TR2010,TPE,MiVaBO-IJCAI2020,cocabo}. Most of them employ non-GP based surrogate models as it is challenging to define a generic kernel over hybrid spaces that can account for complex interactions between variables.

To precisely fill this gap in our knowledge, we propose a novel approch referred as {\em {\bf Hy}brid {\bf B}ayesian {\bf O}ptimization (HyBO)}. HyBO builds GP based surrogate models using diffusion kernels, which are naturally defined over continuous \cite{diffusion_kernel} and discrete spaces \cite{diffusion_kernel_original}. We develop a principled approach to construct diffusion kernels over hybrid spaces. This approach employs the general formulation of additive Gaussian process kernels \cite{additive_gp_kernels} to define {\em additive hybrid diffusion} kernels. The key idea is to assign a base kernel for each discrete/continuous variable and construct an overall kernel by summing over all possible orders of interaction between these kernels. This construction procedure has two advantages: 1) Allows to leverage existing kernels for continuous and discrete spaces; and 2) Can automatically identify the strength of different orders of interaction in a data-driven manner for a given application. . 

A key question about the modeling strength of this hybrid diffusion kernel is whether given sufficient data, can it approximate any black-box function defined over hybrid spaces. This question has been studied in the past in terms of a property called {\em universality} of a kernel \cite{steinwart_univeral,micchelli_universal,sri_univeral,mania_univeral}. We prove that the proposed hybrid diffusion kernel has universal approximation property by composing a known result for continuous diffusion kernels with a novel result for discrete diffusion kernels. Our theoretical results have broader significance going beyond the BO literature.

Our experiments on diverse synthetic benchmarks and real-world applications show that HyBO performs significantly better than state-of-the-art methods. We also empirically demonstrate that superiority of HyBO's performance is due to better surrogate model resulting from the proposed additive hybrid diffusion kernel. 


\noindent {\bf Contributions.} The key contribution of this paper is the development and evaluation of the HyBO approach to perform BO over hybrid spaces. Specific list includes:

\vspace{-1.5ex}

\begin{itemize}
\setlength\itemsep{0em}
    \item Development of a principled approach to construct additive diffusion kernels over hybrid spaces for building GP based surrogate statistical models.
    
    \item Theoretical analysis to prove that additive hybrid diffusion kernel has the universal approximation property.
    
    \item Experiments on synthetic and real-world benchmarks to show that HyBO significantly improves over state-of-the-art methods.  The code and data are available on the GitHub repository \url{https://github.com/aryandeshwal/HyBO}.
    
\end{itemize}

\section{Problem Setup and Hybrid Bayesian Optimization Approach}

\noindent {\bf Problem Setup.} Let $\mathcal{X}$ be a hybrid space to be optimized over, where each element $x \in \mathcal{X}$ is a hybrid structure. Without loss of generality, let each hybrid structure $x$ = $(x_d \in \mathcal{X}_d, x_c \in \mathcal{X}_c) \in \mathcal{X}$ be represented using $m$ discrete variables and $n$ continuous variables, where $x_d$ and $x_c$ stands for the discrete and continuous sub-space of $\mathcal{X}$. Let each discrete variable $v_d$ from $x_d$ take candidate values from a set $C(v_d)$ and each continuous variable $v_c$ from $x_c$ take values from a compact subset of  $\R$. In parts of the ML literature, a distinction is made between categorical and discrete variables based on their values: {\em categorical} refers to an unordered set (e.g., different types of optimizers for neural network training) and {\em discrete} refers to an ordered set ( e.g., number of layers in a neural network). We do not make such distinction because our HyBO approach works for both cases. Concretely, by our definition, a categorical variable is also a discrete variable, i.e., $C(v_d)$ is just the no. of candidate values for categorical variable $v_d$. We are given a space of hybrid structures $\mathcal{X}$. We assume an unknown, expensive real-valued objective function $\mathcal{F}: \mathcal{X} \mapsto \R$, which can evaluate each hybrid structure $x$ (also called an experiment) and produces an output $y$ = $\mathcal{F}(x)$. For example, in high-entropy alloys optimization application, $x_d$ corresponds to the presence/absence of metals and $x_c$ corresponds to their relative concentrations, and $\mathcal{F}(x)$ corresponds to running a physical lab experiment using additive manufacturing techniques. The main goal is to find a hybrid structure $x \in \mathcal{X}$ that approximately optimizes $\mathcal{F}$ by conducting a limited number of evaluations and observing their outcomes.

\noindent {\bf Bayesian Optimization Framework.} BO is a very efficient framework to solve global optimization problems using {\em black-box evaluations of expensive objective functions} \cite{BO-Survey:2016}. BO algorithms  intelligently select the next input for evaluation guided by a learned statistical model to quickly direct the search towards optimal inputs. The three key elements of BO framework are: 

{\em 1) Statistical model} of the true function $\mathcal{F}(x)$. {\em Gaussian Process (GP)} \cite{GP-Book} is the most popular choice for statistical model. GPs allows to incorporate domain knowledge by defining an appropriate kernel over the input space and have better uncertainty quantification ability. A GP over a space $\mathcal{X}$ is a random process from $\mathcal{X}$ to $\R$. It is characterized by a mean function $\mu : \mathcal{X} \mapsto \R$ and a covariance or kernel function $k : \mathcal{X} \times \mathcal{X} \mapsto \R$. 

{\em 2) Acquisition function} ($\mathcal{AF}$) to score the utility of evaluating a candidate input $x \in \mathcal{X}$ based on the statistical model $\mathcal{M}$. Expected improvement (EI) \cite{EI} is a prototypical acquisition function. 

{\em 3)  Optimization procedure} to select the best scoring candidate input for evaluation according to $\mathcal{AF}$. 

\vspace{-1.0ex}
\begin{algorithm}[H]

\caption{HyBO Approach}
\footnotesize
\textbf{Input}: $\mathcal{X}$ = Hybrid input space, 
$\mathcal{K}(x, x')$ = Kernel over hybrid structures,
$\mathcal{AF}(\mathcal{M},x)$ = Acquisition function parametrized by model $\mathcal{M}$ and input $x$,
$\mathcal{F}(x)$ = expensive objective function \\
\textbf{Output}: $\hat{x}_{best}$, the best structure 
\begin{algorithmic}[1]
\STATE Initialize $\mathcal{D}_0 \leftarrow$ initial training data; and $t \leftarrow$ 0
\REPEAT
\STATE Learn statistical model: $\mathcal{M}_t \leftarrow$ \textsc{GP-Learn}($\mathcal{D}_t$, $\mathcal{K}$)
\STATE Compute the next structure to evaluate: \\ $x_{t+1} \leftarrow \mbox{arg}\,max_{x \in \mathcal{X}} \, \mathcal{AF}(\mathcal{M}_t,x)$ \\
\begin{ALC@g}
\STATE $x_c \leftarrow$ Optimize continuous subspace conditioned on assignment to discrete variables $x_d$ 
\STATE $x_d \leftarrow$ Optimize discrete subspace conditioned on assignment to continuous variables $x_c$ 
\end{ALC@g}
\STATE Evaluate objective function $\mathcal{F}(x)$ at $x_{t+1}$ to get $y_{t+1}$
\STATE Aggregate the data: $\mathcal{D}_{t+1} \leftarrow \mathcal{D}_{t} \cup \left\{(x_{t+1}, y_{t+1})\right\}$
\STATE $t \leftarrow t+1$
\UNTIL{convergence or maximum iterations}
\STATE $\hat{x}_{best} \leftarrow \mbox{arg}\,max_{x_t \in \mathcal{D}} \, y_t$
\STATE \textbf{return} the best uncovered hybrid structure $\hat{x}_{best}$
\end{algorithmic}
\label{alg:HyBO}
\end{algorithm}

\vspace{-3.0ex}


\noindent {\bf Hybrid Bayesian Optimization Approach.} Our {\em HyBO} approach is an instantiation of the generic BO framework by instantiating the statistical model and acquisition function optimization procedure for hybrid spaces (see Algorithm~\ref{alg:HyBO}). 


{\em Statistical model over hybrid structures.} We employ GPs to build statistical models. 
To accurately model the complex interactions between discrete and continuous variables, we invoke a principled approach to {\em automatically} construct additive diffusion kernels over hybrid structures by leveraging diffusion kernels over continuous and discrete spaces. 


{\em Acquisition function optimization.} Suppose $\mathcal{M}_t$ is the statistical model at iteration $t$. 
Let us assume that $\mathcal{AF}(\mathcal{M}_t, x)$ is the acquisition function that need to be optimized to select the next hybrid structure $x_{t+1}$ for function evaluation. We solve this problem using an iterative procedure that performs search over continuous sub-space ($x_c$) and discrete sub-space ($x_d$) alternatively. For searching continuous and discrete sub-spaces, we employ CMA-ES \cite{CMA-ES} and hill-climbing with restarts respectively. We observed that {\em one} iteration of optimizing continuous and discrete subspaces gave good results and they were not sensitive to more iterations. All results of HyBO are with one iteration.

\section{Related Work}

The effectiveness of any BO approach over hybrid spaces depends critically on the choice of surrogate model. Prior work explored a variety of surrogate models. SMAC \cite{SMAC} employs random forest, which may suffer from inaccurate uncertainty quantification due to its frequentist estimation. TPE \cite{TPE} models each input dimension {\em independently} by a kernel density estimator, which can be restrictive due to large size of input dimensions and no inter-dependency among models of different input dimensions. MiVaBO \cite{MiVaBO-IJCAI2020} employs a Bayesian linear regressor by defining features that capture the discrete part using BOCS model \cite{BOCS,PSR}, continuous part using random fourier features \cite{RFF}, and pairwise interaction between continuous and discrete features. As the number of parameters increase, it will need a lot of training examples for learning accurate statistical model. 

GP based models overcome the drawbacks of all the above methods. \cite{lobato} provided a solution for BO over discrete spaces using an input-transformed kernel. A recent work referred as CoCaBO \cite{cocabo} employs a sum kernel (summing a Hamming kernel over discrete subspace and a RBF kernel over continuous subspace) to learn GP models and showed good results over SMAC and TPE. Unfortunately, the sum kernel captures limited interactions between discrete and continuous variables. In contrast, our additive hybrid diffusion kernel allows to capture higher-order interactions among hybrid variables and our data-driven approach can automatically learn the strengths of these interactions from training data. HyperBand (HB) \cite{HB} and its model-based variant BOHB \cite{BOHB} are efficient {\em multi-fidelity methods} for hyper-parameter optimization that build on existing methods to optimize hybrid spaces. Our HyBO approach is complementary to this line of work.

Prior methods perform search over discrete and continuous subspaces (e.g., gradient descent) to solve the acquisition function optimization problem. SMAC employs a {\em hand-designed} local search procedure. MiVaBO uses integer program solvers to search discrete subspace. Learning methods to improve the accuracy of search \cite{L2S-DISCO} are complementary to SMAC, MiVABO, and HyBO. CoCaBO maintains a separate multi-armed bandit for each discrete variable and employs the EXP3 algorithm \cite{EXP3} to select their values {\em independently}. This method does not exploit dependencies among variables, which can be detrimental to accuracy. TPE samples from the learned density estimator to pick the best input for evaluation.

\section{Diffusion Kernels over Hybrid Structures}
\label{sec:hybrid-kernels}

We first provide the details of key mathematical and computational tools that are needed to construct hybrid diffusion kernels. Next, we describe the algorithm to automatically construct additive diffusion kernels over hybrid structures. Finally, we present theoretical analysis to show that hybrid diffusion kernels satisfy universal approximation property.  

\subsection{Key Mathematical and Computational Tools}\label{sec:key_math}

Diffusion kernels \cite{diffusion_kernel,diffusion_statistical_manifold} are inspired from the diffusion processes occurring in physical systems like heat and gases. The mathematical formulation of these processes naturally lends to kernels over both continuous and discrete spaces(e.g., sequences, trees, and graphs). 


\noindent {\bf Diffusion kernel over continuous spaces.} The popular radial basis function (RBF) kernel (also known as Gaussian kernel) \cite{diffusion_kernel} is defined as follows:
\begin{align}
k(x, x') = \frac{1}{2 \pi \sigma^2} e^{-\|x-x'\|^2/2\sigma^2} \label{eqn:rbf_kernel}  
\end{align}
 where $\sigma$ is the length scale hyper-parameter. This is the solution of the below continuous diffusion (heat) equation:
\begin{align}
\frac{\partial}{\partial t}k_{x_0} (x, t) = \Delta k_{x_0} (x, t) \label{eqn:continuous_heat_equation}
\end{align} 
where $\Delta$ = $\frac{\partial^2}{\partial x_1^2} + \frac{\partial^2}{\partial x_2^2} \cdots \frac{\partial^2}{\partial x_D^2}$ is the second-order differential operator known as the {\em Laplacian operator}, and $k_{x_0} (x, t)$ = $k(x, x')$ with $x'$ = $x_0$ and $t$ = $\sigma^2/2$.


\subsection{Diffusion Kernel over discrete spaces}
The idea of diffusion kernels for continuous spaces is extended to discrete structures (e.g., sequences, graphs) \cite{diffusion_kernel_original} by utilizing the spectral properties of a graph representation of the discrete space. 
A discrete analogue of the Equation \ref{eqn:continuous_heat_equation} can be constructed by employing the matrix exponential of a graph and the {\em graph Laplacian operator} $L$ as given below:
\begin{align}
\frac{\partial}{\partial \beta} e^{\beta L} = L e^{\beta L}					\label{eqn:discrete_diffusion_kernel}
\end{align}
where $L$ is the graph Laplacian of a suitable graph representation of the  discrete input space and $\beta$ is a hyper-parameter of the resulting diffusion kernel similar to the length scale parameter $\sigma$ of the RBF kernel. The solution of Equation \ref{eqn:discrete_diffusion_kernel} defines a positive-definite kernel for discrete spaces known as the discrete diffusion kernel. 

According to Equation \ref{eqn:discrete_diffusion_kernel}, one important ingredient required for defining diffusion kernels on discrete spaces is a suitable graph representation for discrete spaces. One such representation was proposed in a recent work \cite{combo}. In this case,   
the entire discrete space is represented by a combinatorial graph $G$. Each node in the vertex set $V$ of the graph corresponds to one candidate assignment of all the discrete variables. Two nodes are connected by an edge if the Hamming distance between the corresponding assignments for all discrete variables is exactly one. The diffusion kernel over this representation is defined as follows:
\begin{align}
    k(V, V) &= \exp(-\beta L(G)) \label{eqn:main_discrete_kernel}\\ 
    k(V, V) &= \Phi \exp(-\beta \Pi) \Phi^T
\end{align}
where $\Phi$ = $[\phi_1, \cdots, \phi_{|V|}]$ is the eigenvector matrix  and $\Pi$ = $[\pi_1, \cdots, \pi_{|V|}]$ is the eigenvalue matrix, where $\phi_i$'s and $\pi_i$'s are the eigenvectors and eigenvalues of the graph Laplacian $L(G)$ respectively. Although this graph representation contains an exponential number of nodes, \cite{combo} computes the graph Laplacian $L(G)$ by decomposing it over the Cartesian product ($\diamond$) of $m$ (number of discrete variables) sub-graphs ($G_1, G_2 \cdots, G_m$) with each sub-graph $G_i$ representing one variable individually.
This algorithmic approach has time-complexity $O(\sum_{i=1}^m (C(v_i))^3)$, where $C(v_i)$ is the number of candidate values (arity) for the $i$th discrete variable. However, this method is computationally expensive, especially, for problems with large-sized arity.

To avoid this computational challenge, we leverage prior observation in \cite{diffusion_kernel_original} which provides a {\em closed-form} of the discrete diffusion kernel by exploiting the structure of the above combinatorial graph representation. We explain this observation for binary variables $\{0, 1\}$. From its definition in Equation \ref{eqn:main_discrete_kernel}, the discrete diffusion kernel over single-dimensional input will be:

\begin{align}\label{eqn:binary_dd_closed_form}
    k(x_d, x_d') =  \left\{
	\begin{array}{ll}
		(1 - e^{-2\beta})   & \mbox{if } x_d \neq x_d' \\
		(1 + e^{-2\beta}) & \mbox{if } x_d = x_d' 
	\end{array}
\right. 
\end{align}
Since the kernel over $m > 1$ dimensions is defined using the Kronecker product over $m$ dimensions, the above expression is easily extended to multiple dimensions setting giving:

\begin{align}
k(x_d, x_d') = \prod_{i=1}^m \frac{(1 - e^{-2\beta_i})}{(1 + e^{-2\beta_i})}^{\delta(x_d^i, x_d'^i)}
\end{align}
where $\delta(x_d^i, x_d'^i)$ = $0$ if $x_d^i$ is equal to $x_d'^i$ and $1$ otherwise. The subscript $d$ denotes that the variables are discrete and the superscript refers to the $i$th dimension of the discrete subspace. For general (discrete spaces with arbitray categories), we follow the same observation \cite{diffusion_kernel_original} and use the following constant-time expression of the discrete diffusion kernel in our method:
\begin{align}
    k(x_d, x_d') = \prod_{i=1}^m \left( \frac{1-e^{-C(v_i)\beta_i}}{1 + (C(v_i)-1) e^{-C(v_i)\beta_i}} \right)^{\delta(x_d^i, x_d'^i)} \label{eqn:final_discrete_diffusion}
\end{align}

\subsection{Diffusion Kernels over Hybrid Spaces} 


\noindent {\bf Unifying view of diffusion kernels.} Our choice of diffusion kernels is motivated by the fact that they can be naturally defined for both discrete and continuous spaces. In fact, there is a nice transition of the diffusion kernel from discrete to continuous space achieved by continuous space limit operation. More generally, both discrete and continuous diffusion kernel can be seen as continuous limit operation on two parameters of random walks: {\em time} and {\em space}. For illustration, consider a random walk on an evenly spaced grid where mean time of jump is $t$ and mean gap between two points is $s$. If $t \to 0$, the resulting continuous time and discrete space random walk generates the diffusion kernel on discrete spaces. Additionally, in the limit of the grid spacing $s$ going to zero, the kernel will approach the continuous diffusion kernel. 


\noindent {\bf Algorithm to construct hybrid diffusion kernels.} We exploit the general formulation of additive Gaussian process kernels \cite{additive_gp_kernels} to define an {\em additive hybrid diffusion} kernel over hybrid spaces. The key idea is to assign a base kernel {\em for each input dimension $i \in \{1, 2, \cdots, m+n\}$}, where $m$ and $n$ stand for the number of discrete and continuous variables in hybrid space $\mathcal{X}$; and construct an overall kernel by summing all possible orders of interactions (upto $m+n$) between these base kernels. In our case, the RBF kernel and the discrete diffusion kernel acts as the base kernel for continuous and discrete input dimensions respectively. The $p^{th}$ order of interaction (called {\em $p^{th}$ additive kernel}) is defined as given below:
\begin{align*}
    \mathcal{K}_{p} = \theta_p^2 \sum_{1\leq i_1 < i_2 < \cdots, i_p\leq m+n} \left(\prod_{d=1}^p   k_{i_d}(x_{i_d}, x_{i_d}')\right) \label{eqn:additive_kernel_def}
\end{align*}
where $\theta_p$ is a hyper-parameter associated with each additive kernel and $k_{i_d}$ is the base kernel for the input dimension $i_d$. In words, the $p$th additive kernel is a sum of $m+n \choose p$ terms, where each term is a product of $p$ distinct base kernels. Estimation of $\theta_p$ hyper-parameter from data allows automatic identification of important orders of interaction for a given application.  The overall {\em additive hybrid diffusion kernel} $\mathcal{K}_{HYB}(x, x')$ over hybrid spaces is defined as the sum of all orders of interactions as given below:
\begin{align}
    \mathcal{K}_{HYB} &= \sum_{p=1}^{m+n} \mathcal{K}_{p}  \\
    \mathcal{K}_{HYB} &= \sum_{p=1}^{m+n} (\theta_p^2 \sum_{i_1, \cdots, i_p} \prod_{d=1}^p   k_{i_d}(x_{i_d}, x_{i_d}')) \label{eqn:additive_diff_kernel}
\end{align}
It should be noted that the RHS in Equation \ref{eqn:additive_diff_kernel} requires computing a sum over exponential number of terms. However, this sum can be computed in polynomial time using Newton-Girard formula for elementary symmetric polynomials \cite{additive_gp_kernels}. It is an efficient formula to compute the $p^{th}$ additive kernel recursively as given below:
\begin{align}
    \mathcal{K}_{p} = \theta_p^2 \cdot \left(\frac{1}{p} \sum_{j=1}^p (-1)^{(j-1)} \mathcal{K}_{p-j} S_j\right)
\end{align}
where $S_j = \sum_{i=1}^{m+n} k_i^{j}$  is the $j$th power sum of all base kernels $k_j$ and the base case for the recursion can be taken as 1 (i.e., $\mathcal{K}_0 = 1$). This recursive algorithm for computing additive hybrid diffusion kernel has the time complexity of $\mathcal{O}((n+m)^2)$.


\noindent {\bf Data-driven specialization of kernel for a given application.} In real-world applications, the importance of different orders of interaction can vary for optimizing the overall performance of BO approach (i.e., minimizing the number of expensive function evaluations to uncover high-quality hybrid structures). For example, in some applications, we  may not require all orders of interactions and only few will suffice. The $\theta_p$ hyper-parameters in the additive hybrid diffusion kernel formulation allows us to identify the strength/contribution of the $p$th order of interaction for a given application in a {\em data-driven} manner. We can compute these parameters (along with the hyper-parameters for each base kernel) by maximizing the marginal log-likelihood, but we consider a fully Bayesian treatment by defining a prior distribution for each of them. This is important to account for the uncertainty of the hyper-parameters across BO iterations. The acquisition function $\mathcal{AF}(x)$ is computed by marginalizing the hyper-parameters as given below:
\begin{align}
    \mathcal{AF}(x; \mathcal{D}) = \int \mathcal{AF}(x; D, \Theta) p(\Theta|D) d\Theta \label{acquisition}
\end{align}
where $\Theta$ is a variable representing all the hyperparameters ($\sigma$ for continuous diffusion kernel, $\beta$ for discrete diffusion kernel, and $\theta$ for strengths of different orders of interaction in hybrid diffusion kernel) and $\mathcal{D}$ represents the aggregate dataset containing the hybrid structure and function evaluation pairs. The posterior distribution over the hyper-parameters is computed using slice sampling \cite{slice_sampling}. 


\vspace{-1.0ex}

\subsection{Theoretical Analysis}\label{sec:theoretical}
Intuitively, a natural question to ask about the modeling power of a kernel is whether (given enough data) it can approximate (with respect to a suitable metric) any black-box function defined over hybrid spaces. This is a minimum requirement that should guide our choice of kernel in the given problem setting. This question has been studied widely in the form of a key property called {\em universality} of a kernel \cite{steinwart_univeral,micchelli_universal,sri_univeral,mania_univeral}. In this section, we prove the universality of the {\em additive hybrid diffusion kernel} by combining the existing result on the universality of RBF (Gaussian) kernel with a novel result proving the universality of discrete diffusion kernels.

\vspace{-1.0ex}

\begin{restatable}{proposition}{cl}
\cite{steinwart_univeral,micchelli_universal} Let $\mathcal{X}_c$ be a compact and non-empty subset of $\R^n$. The RBF kernel in Equation \ref{eqn:rbf_kernel} is a universal kernel on $\mathcal{X}_c$.\label{theorem:universal_cont} 
\end{restatable}

\noindent A kernel $k$ defined on an input space $\mathcal{X}_c$ has a unique correspondence with an associated Reproducing Kernel Hilbert Space (RKHS) of functions $\mathcal{H}_k$ defined on $\mathcal{X}_c$ \cite{svm_book}. For compact metric input spaces $\mathcal{X}_c$, a kernel $k$ is called universal if the RKHS $\mathcal{H}_k$ defined by it is dense in the space of continuous functions $C(\mathcal{X}_c)$. \cite{steinwart_univeral} proved the universality of the RBF (Gaussian) kernel with respect to the uniform norm. \cite{micchelli_universal} established universality for a larger class of translation invariant kernels. \cite{sri_univeral} discussed various notions of universality and connected to the concept of {\em characteristic kernels}.

\vspace{-1.5ex}

\begin{restatable}{proposition}{dl}
Let $\mathcal{X}_d$ be the discrete space $\{0, 1\}^m$ and a psuedo-boolean function on $\mathcal{X}_d$ is defined as $f : \mathcal{X}_d \mapsto \R$. 
The discrete diffusion kernel is a universal kernel on $\mathcal{X}_d$. \label{theorem:universal_discrete}
\end{restatable}

{\noindent \bf Proof.} A Reproducing Kernel Hilbert Space $\mathcal{H}_k$ associated with a kernel $k: \mathcal{X} \times \mathcal{X} \mapsto \mathbb{R}$ is defined as:
\begin{align}
    \mathcal{H}_k = cl(span\{k(x, \cdot), \forall x \in \mathcal{X}\})
\end{align}
where $cl$ represents the closure and $k(x, \cdot)$ is called as the feature map of $x$ \cite{svm_book}.

In our setting, a kernel $k$ defined on discrete input space $\mathcal{X}_d$ is universal if and only if any pseudo-Boolean function $f$ can be written as a linear combination  of functions ($k(x_{i_d}, \cdot), \forall x_{i_d} \in \mathcal{X}_d$) in the RKHS $\mathcal{H}_k$ \cite{mania_univeral,gretton_universal}, i.e.
\begin{align} \forall f : \mathcal{X}_d \mapsto \R; \quad
\exists a_i \in \R; 
    f = \sum_i a_i k(x_{i_d}, \cdot); 
\end{align}

We prove that this is true by computing the explicit form of functions ($k(x_{i_d}, \cdot), \forall x_{i_d} \in \mathcal{X}_d$) existing in the RKHS $\mathcal{H}_k$ of the discrete diffusion kernel. To see this, we exploit the structure of the combinatorial graph representation of the discrete space discussed in Section \ref{sec:key_math}. The discrete diffusion kernel is defined in terms of the eigenvectors $\phi_i$ and eigenvalues $\pi_i$ of the graph  Laplacian $L(G)$ as follows:
\begin{align}
    k(x_d, x_d') &= \sum_{i=1}^{2^n} \phi_i[x_d] \exp(-\beta \pi_i) \phi_i[x_d'] \label{eqn:kernel} 
\end{align}
Since the combinatorial graph $G$ is generated by the Cartesian product over sub-graphs $G_i$ (one for each discrete variable), the eigenvectors term $\phi_i[x_d]$  can be calculated via an explicit formula, i.e., $\phi_i[x_d] = -1^{w^Tx_d}$, where $w$ is a binary vector of size $n$ \cite{chung1997spectral} (number of discrete variables). 
\begin{align}
    k(x_d, x_d') &= \sum_{i=1}^{2^n} -1^{w^Tx_d}  \exp(-\beta \pi_i) -1^{w^Tx_d'} \\
<k(x_d, \cdot), & k(x_d', \cdot)> =  \sum_{i=1}^{2^n} -1^{w^Tx_d}  \exp(-\beta \pi_i) -1^{w^Tx_d'}
\end{align}
where the inner product in LHS follows from the reproducing property \cite{svm_book} of a kernel $k$. Therefore, the functions $k(x_d, \cdot)$ in the RKHS  $\mathcal{H}_k$ of the discrete diffusion kernel are of the form $\{-1^{{w_j}^Tx_d} ; w_j \in [0, 2^n-1]\}$, which is the well-known {\em Walsh Basis} \cite{walsh_basis} for pseudo-Boolean functions. Therefore, any pseudo-Boolean function $f$ can be represented by a linear combination of functions in $\mathcal{H}_k$ since they form a basis.

\vspace{-1.5ex}

\begin{restatable}{theorem}{mt}
Let $\mathcal{X}_c$ be a compact and non-empty subset of $\R^n$ and $\kappa_c$ be  RBF kernel on $\mathcal{X}_c$. Let $\mathcal{X}_d$ be the discrete space $\{0, 1\}^m$ and $\kappa_d$ be discrete diffusion kernel on $\mathcal{X}_d$. The additive hybrid diffusion kernel defined in Eqn \ref{eqn:additive_diff_kernel}, instantiated with $k_c$ and $k_d$ for continuous and discrete spaces respectively, is a universal kernel for the hybrid space $\mathcal{X}_c \times \mathcal{X}_d$. \label{theorem:universal_hybrid} 
\end{restatable}
\noindent According to Equation \ref{eqn:additive_kernel_def}, any $p$th order of interaction term in the additive hybrid diffusion kernel is defined as $\left(\prod_{d=1}^p   k_{i_d}(x_{i_d}, x_{i_d}')\right)$. Therefore, if each $k_{i_d}$ is universal over its corresponding dimension $X_{i_d}$ (which is true from Propositions 1 and 2), we need to show that the product $\left(\prod_{d=1}^p   k_{i_d}(x_{i_d}, x_{i_d}')\right)$ is universal over the union of dimensions $\mathcal{X}_{i_1} \times \mathcal{X}_{i_2} \cdots \times \mathcal{X}_{i_p}$. This was proven by Lemma A.5 in \cite{heirarchical_steinwart}. We provide the lemma here for completeness.
\begin{lemma}{From \cite{heirarchical_steinwart}}\label{tensor-universal}
Let $\mathcal{X}\subset \R^m$ be a compact and non-empty subset, $I,J\subset \{1,\dots,m\}$ be non-empty,
and $k_I$ and $k_J$ be universal kernels on $\mathcal{X}_I \times \mathcal{X}_J$, respectively. Then 
$k_I\otimes k_J$ defined by 
\begin{displaymath}
k_I\otimes k_J(x , x' ) := k_I(x_I, x'_I) \cdot k_J(x_J, x'_J)   
\end{displaymath}
for all  $x , x' \in \mathcal{X}_{I} \times  \mathcal{X}_{J}$
  is a universal kernel on $\mathcal{X}_{I} \times  \mathcal{X}_{J}$.
\end{lemma}

Since both continuous and discrete spaces are compact and Lemma \ref{tensor-universal} is true for arbitrary compact spaces, each order of interaction is universal with respect to its corresponding ambient dimension $\mathcal{X}_{i_1} \times \mathcal{X}_{i_2} \cdots \times \mathcal{X}_{i_p}$. In particular, it is true for $m+n$th order of interaction which is defined over the entire hybrid space $\mathcal{X}_c \times \mathcal{X}_d$ which proves the theorem.

\vspace{-1.0ex}

\section{Experiments and Results}

We first describe our experimental setup. Next, we discuss experimental results along different dimensions.

\vspace{-1.0ex}

\subsection{Benchmark Domains}

\vspace{0.5ex}

\noindent {\bf Synthetic benchmark suite.} \texttt{bbox-mixint} is a challenging mixed-integer blackbox optimization benchmark suite \cite{bbox_mixint} that contains problems of varying difficulty. 
This benchmark suite is available via COCO platform\footnote{\url{https://github.com/numbbo/coco}}. We ran experiments with multiple problems from this benchmark, but for brevity, we present canonical results on four benchmarks (shown in Table \ref{tab:bbox}) noting that all the results show similar trends. 


\begin{table}[t!]
\centering
\begin{tabular}{|l | c | c |}  
\hline
Name &  Name in the suite & Dimension \\
\hline
\hline
Function 1  & f001\_i01\_d10  &  10 (8d, 2c)  \\
Function 2 & f001\_i02\_d10 & 10 (8d, 2c) \\
Function 3 & f001\_i01\_d20 & 20 (16d, 4c)  \\
Function 4 & f001\_i02\_d20 & 20 (16d, 4c)  \\
\hline
\end{tabular}
\vspace{-1ex}
\caption{Benchmark problems from bbox-mixint suite.} 
\label{tab:bbox}
\vspace{-1ex}
\end{table}

{\bf \noindent Real world benchmarks.} We employ six diverse real-world domains. The complete details (function definition, bounds for input variables etc.) 
are in the Appendix. 

\vspace{-0.5ex}

{\bf 1) Pressure vessel design optimization.} This mechanical design problem \cite{pressure_vessel_1,pressure_vessel_2} involves minimizing the total cost of a cylindrical pressure vessel. There are two discrete (thickness of shell and head of pressure vessel) and two continuous (inner radius and length of cylindrical section) variables.

\vspace{-0.5ex}

{\bf  2) Welded beam design optimization.}  The goal in this material engineering domain \cite{weld_design_1,weld_design_2} is to design a welded beam while minimizing the overall cost of the fabrication. There are six variables: two discrete (type of welding configuration and bulk material of the beam) and four continuous (weld thickness, welded joint length, beam width and thickness). 

\vspace{-0.5ex}

{\bf 3) Speed reducer design optimization.} In this domain from NASA \cite{speed_reducer}, the goal is to minimize the weight of a speed reducer defined over seven input variables: one discrete (number of teeth on pinion) and six continuous (face width, teeth module, lengths of shafts between bearings, and diameters of the shafts)

\vspace{-0.5ex}

{\bf 4) Optimizing control for robot pushing.} This is a 14 dimensional control parameter tuning problem, where a robot is trying to push objects toward a goal location \cite{EBO}. We consider a hybrid version of this problem by discretizing  ten input variables corresponding to location of the robot and number of simulation steps. The remaining four parameters corresponding to rotation are kept as continuous.

\vspace{-0.5ex}

{\bf 5) Calibration of environmental model.} The problem of calibration and uncertainty analysis of expensive environmental models is very important in scientific domains \cite{em_func,astudillo2019bayesian}. There are four input variables (one discrete and three continuous).

\begin{figure*}[h!]
\centering
\subfloat[Subfigure 2 list of figures text][]{
\includegraphics[width=0.25\textwidth]{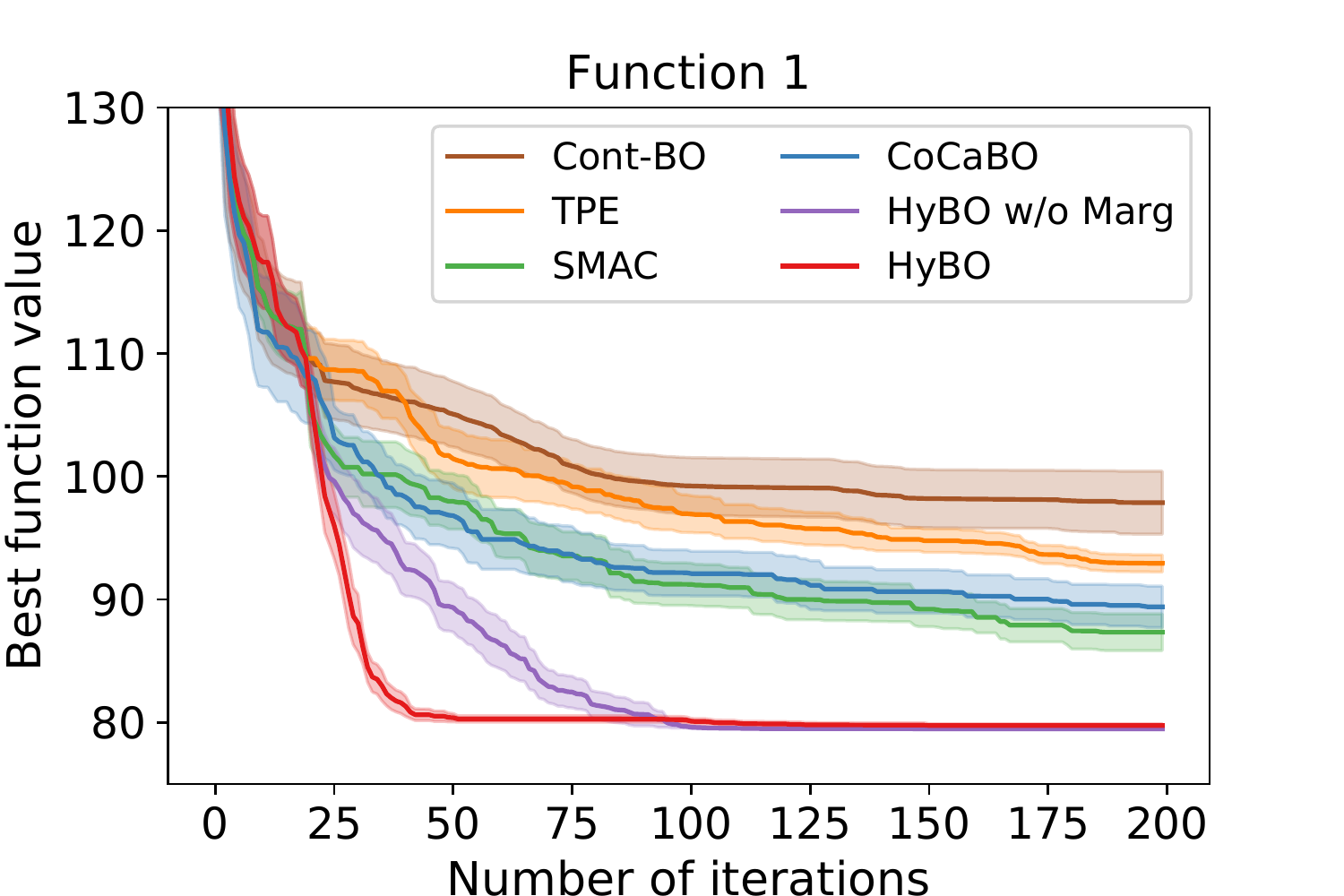}
\label{fig:function_1}}
\subfloat[Subfigure 1 list of figures text][]{
\includegraphics[width=0.25\textwidth]{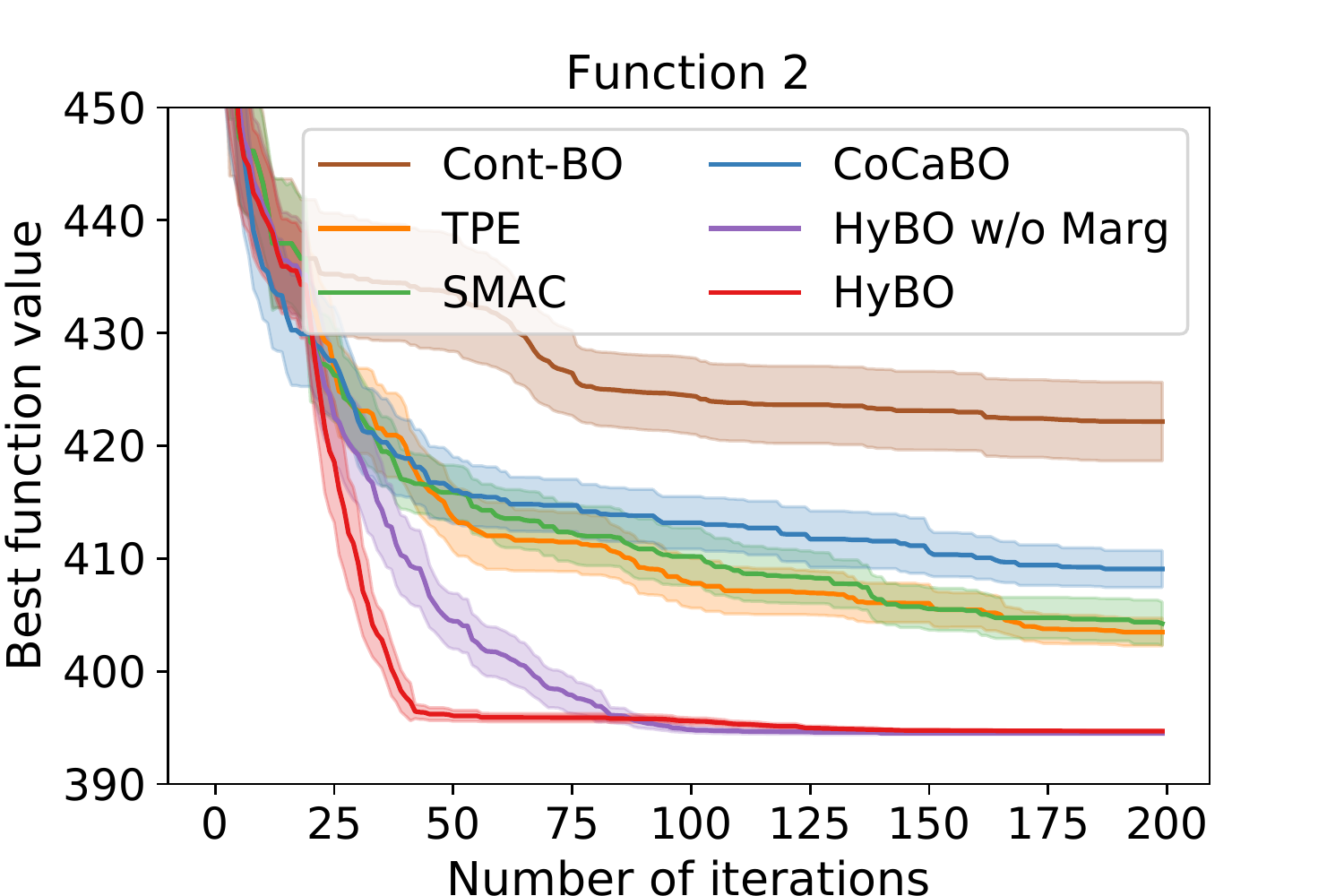}
\label{fig:function_2}}
\subfloat[Subfigure 2 list of figures text][]{
\includegraphics[width=0.25\textwidth]{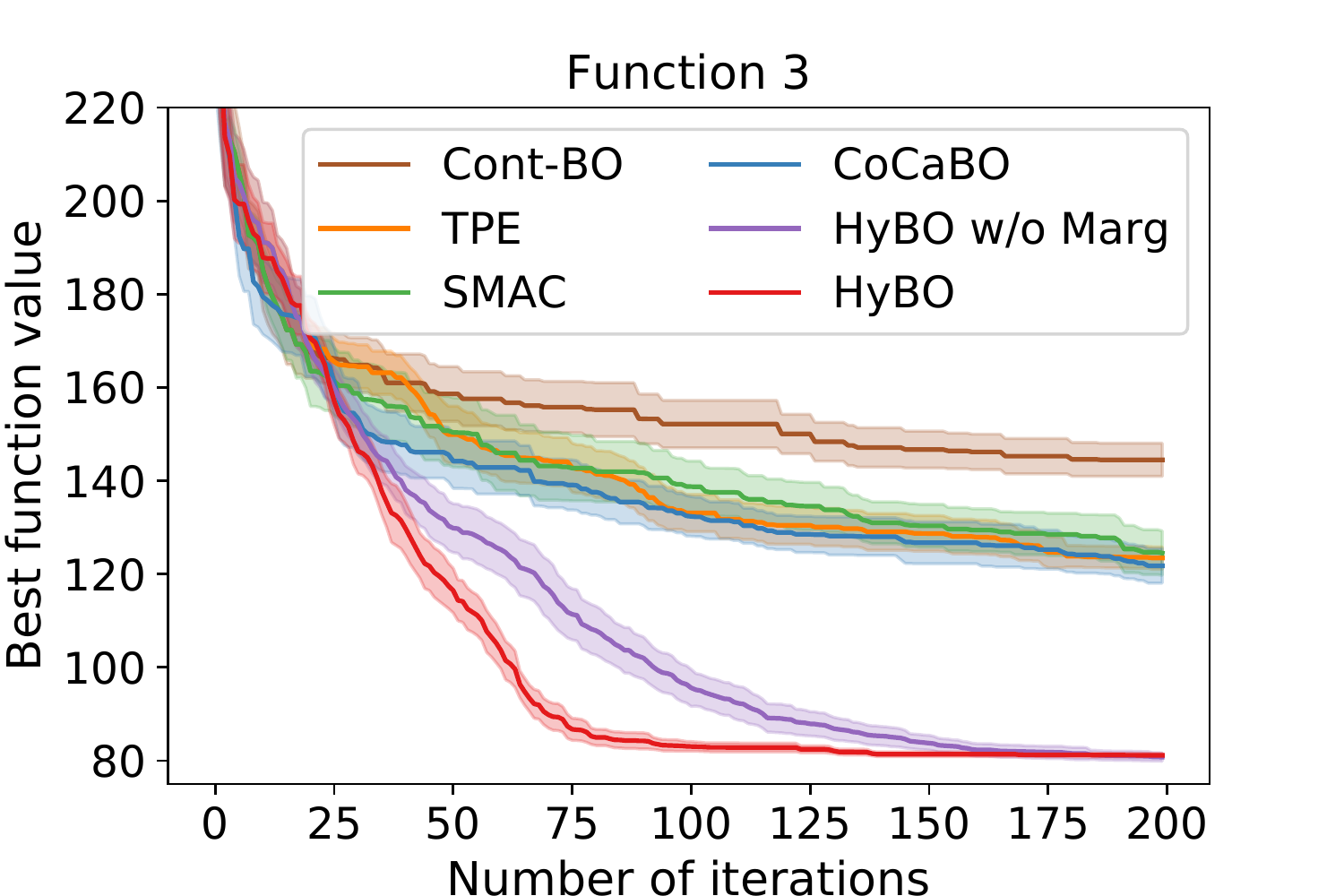}
\label{fig:function_3}}
\subfloat[Subfigure 1 list of figures text][]{
\includegraphics[width=0.25\textwidth]{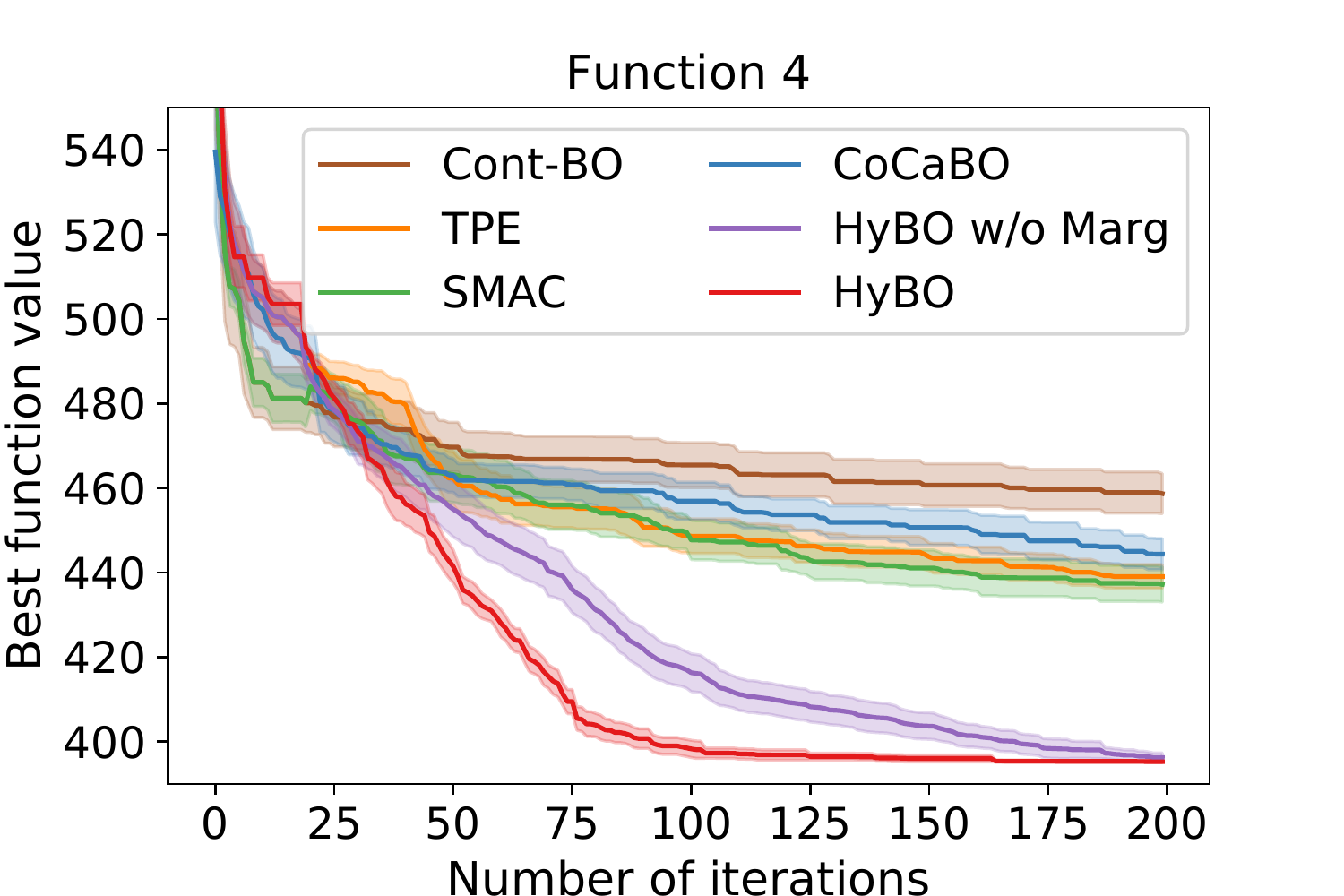}
\label{fig:function_4}}
\caption{Results of HyBO and state-of-the-art baselines on bbob-mixint benchmark suite for functions shown in Table \ref{tab:bbox}.} 
\label{fig:synthetic}
\end{figure*}

\vspace{-0.5ex}

{\bf 6) Hyper-parameter optimization.} We consider hyper-parameter tuning of a neural network model on a diverse set of benchmarks \cite{nn_hpo}: five discrete (hidden layer size, activation type, batch size, type of learning rate, and whether to use early stopping or not) and three continuous (learning rate initialization, momentum parameter, and regularization coefficient) hyper-parameters.

\vspace{-1.0ex}

\subsection{Experimental Setup}

\noindent {\bf Baseline methods.} We compare HyBO with four strong baselines: 1) \texttt{CoCaBO}, a state-of-the-art method \cite{cocabo}; 2) \texttt{SMAC} \cite{SMAC}; 3) \texttt{TPE} \cite{TPE}; 4) \texttt{HyBO w/o Marg} is a special case of HyBO, where we do not perform marginalization over the hyper-parameters of the hybrid diffusion kernel; 
and 5) \texttt{Cont-BO} treats discrete variables as continuous and performs standard BO over continuous spaces (both modeling and acquisition function optimization). We did not include \texttt{MiVaBO} \cite{MiVaBO-IJCAI2020} as there was no publicly available implementation \cite{PC:2020} \footnote{Personal communication with the lead author.}.



\noindent {\bf Configuration of algorithms and baselines.} We configure \texttt{HyBO} as follows. We employ uniform prior for the length scale hyperparameter ($\sigma$) of the RBF kernel. Horse-shoe prior is used for $\beta$ hyper-parameter of the discrete diffusion kernel (Equation \ref{eqn:final_discrete_diffusion}) and hyper-parameters $\theta$ of the additive diffusion kernel (Equation \ref{eqn:additive_kernel_def}). We employ expected improvement \cite{EI} as the acquisition function. 
For acquisition function optimization, we perform iterative search over continuous and discrete sub-spaces as shown in Algorithm~\ref{alg:HyBO}. For optimizing discrete subspace, we run local search with 20 restarts. We normalize each continuous variable to be in the range $[-1, 1]$  and employed CMA-ES algorithm \footnote{\url{https://github.com/CMA-ES/pycma}} for optimizing the continuous subspace. We found that the results obtained by CMA-ES were not sensitive to its hyper-parameters. Specifically, we fixed the population size to 50 and initial standard deviation to 0.1 in all our experiments.
We employed the open-source python implementation of 
CoCaBO \footnote{\url{https://github.com/rubinxin/CoCaBO_code}}, SMAC \footnote{\url{https://github.com/automl/SMAC3}}, and TPE \footnote{\url{https://github.com/hyperopt/hyperopt}}.

All the methods are initialized with same random hybrid structures. We replicated all experiments for 25 different random seeds and report the mean and two times the standard error in all our figures. 


\noindent {\bf Evaluation metric.} We use the best function value achieved after a given number of iterations (function evaluations)  as a metric to evaluate all methods. The method that uncovers high-performing hybrid structures with less number of function evaluations is considered better. 

\vspace{-1.0ex}

\subsection{Results and Discussion}

\vspace{1.5ex}

\begin{figure*}[h!]
\centering
\subfloat[Subfigure 2 list of figures text][]{
\includegraphics[width=0.25\textwidth]{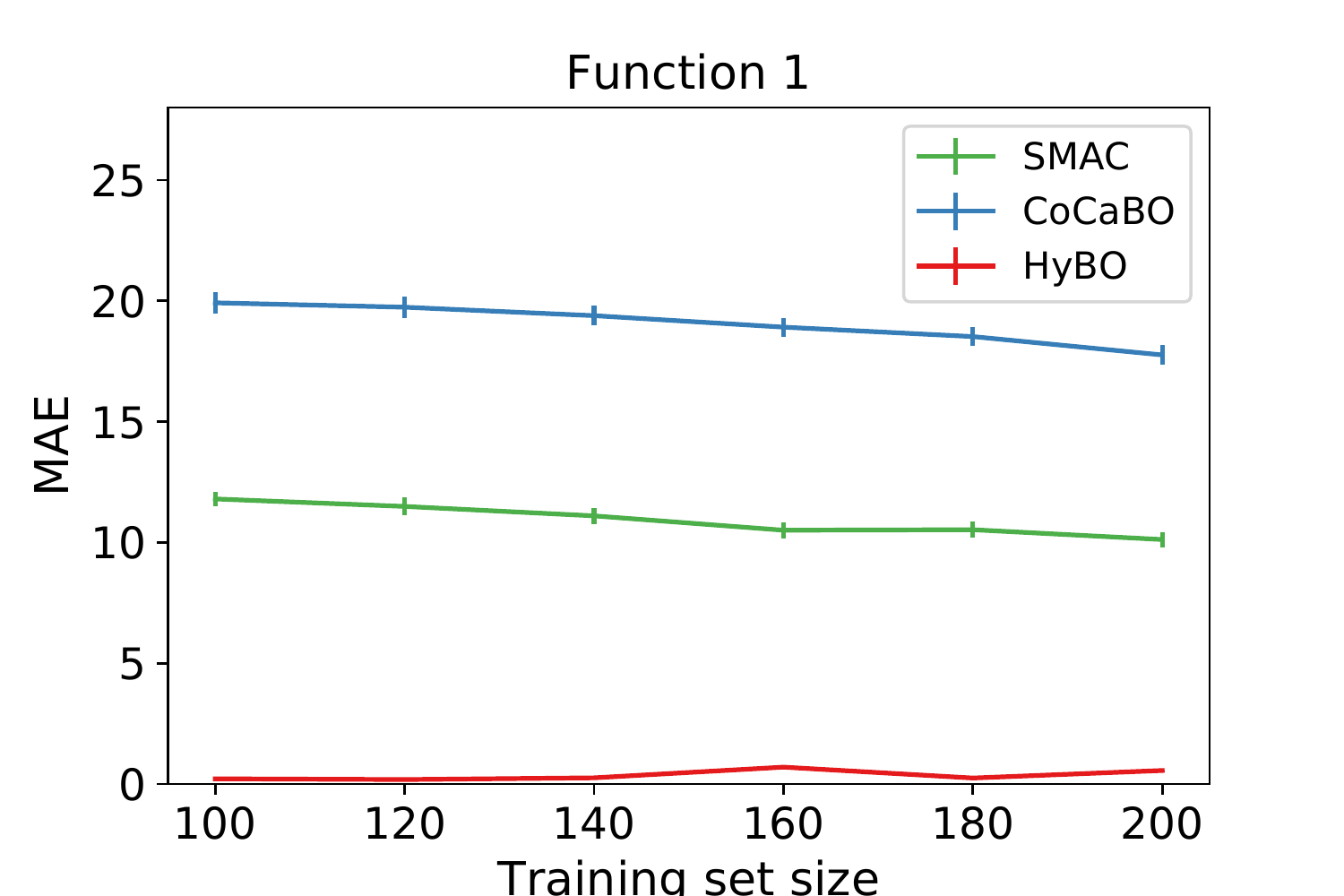}
\label{fig:mae_function_1}}
\subfloat[Subfigure 1 list of figures text][]{
\includegraphics[width=0.25\textwidth]{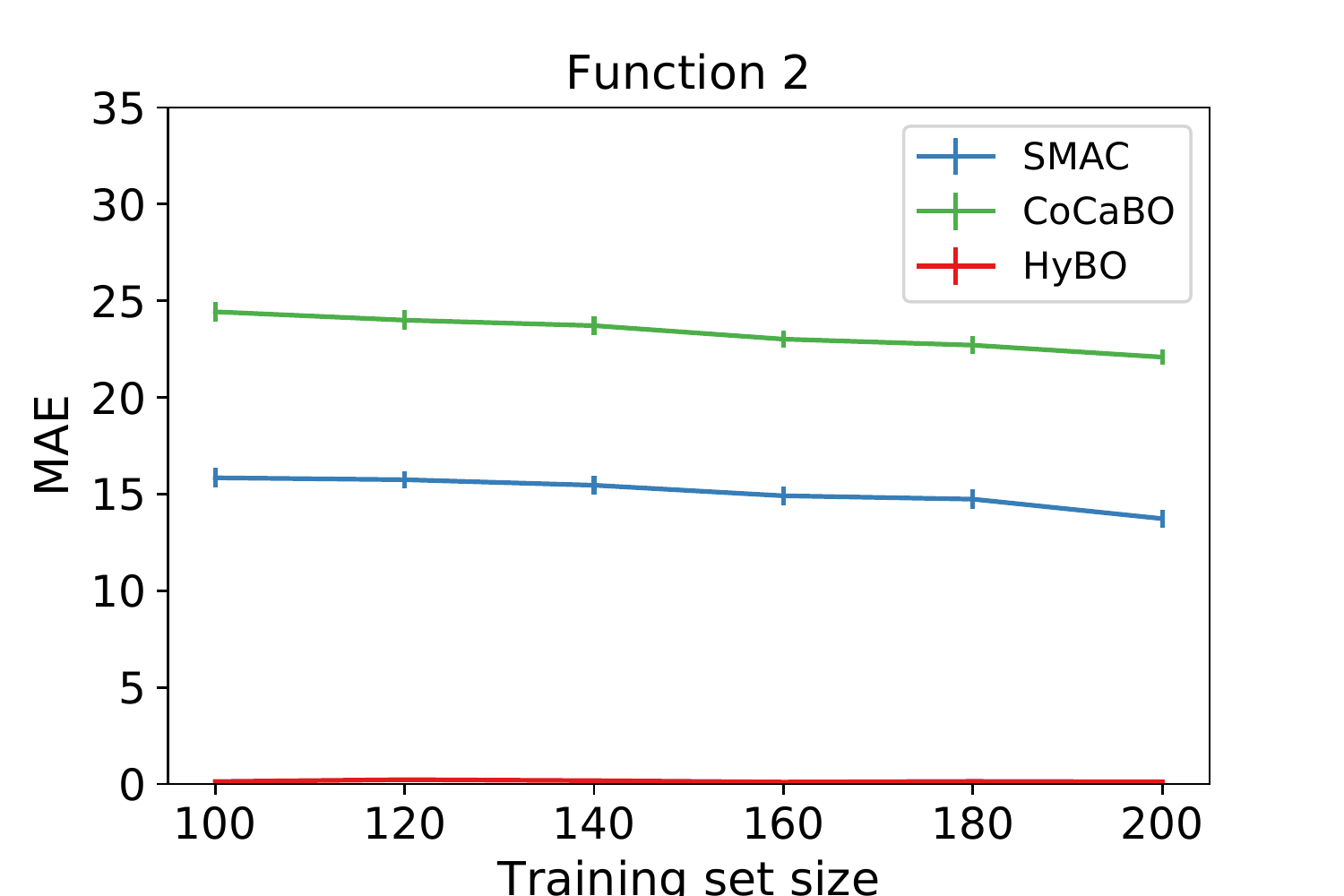}
\label{fig:mae_function_2}}
\subfloat[Subfigure 2 list of figures text][]{
\includegraphics[width=0.25\textwidth]{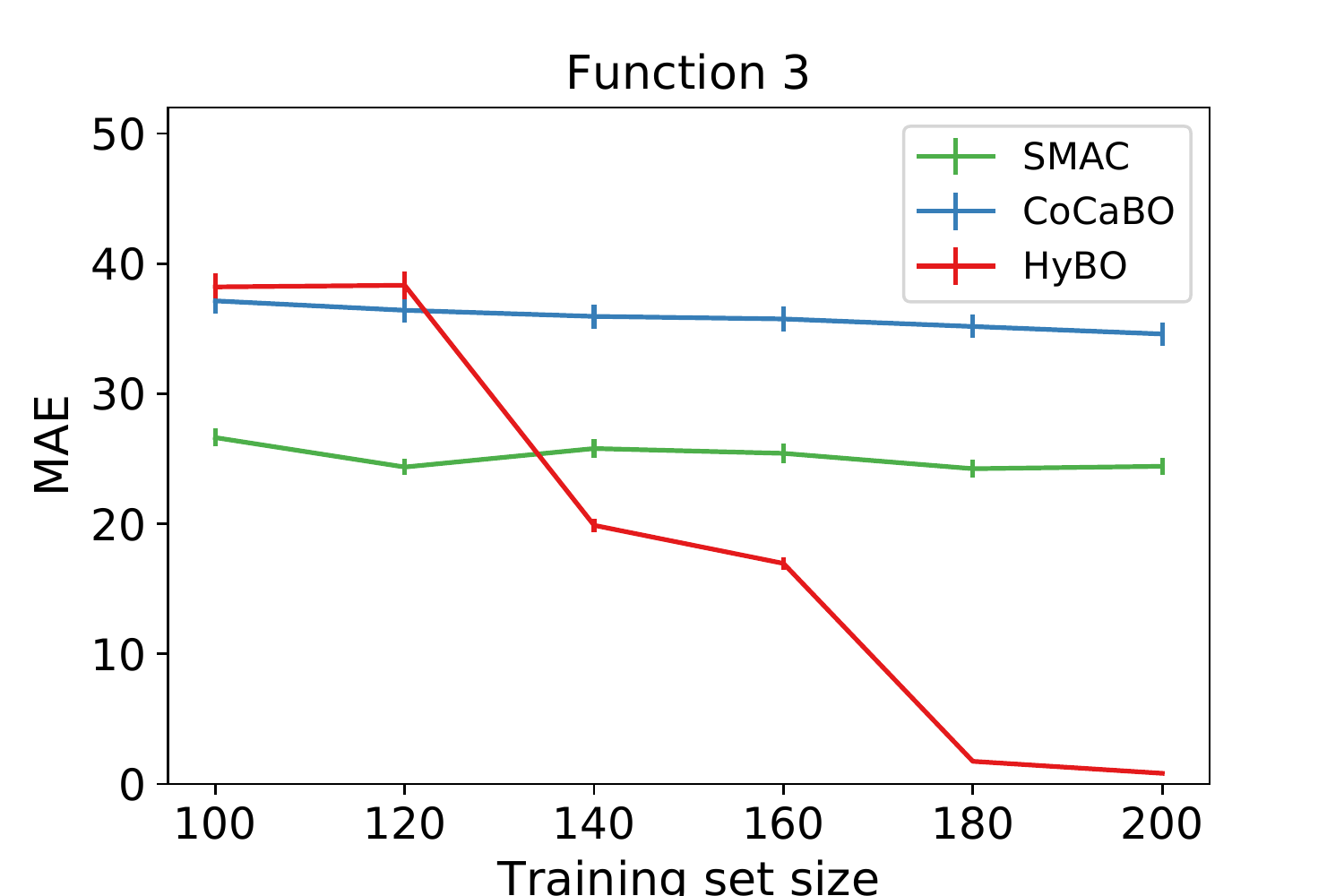}
\label{fig:mae_function_3}}
\subfloat[Subfigure 2 list of figures text][]{
\includegraphics[width=0.25\textwidth]{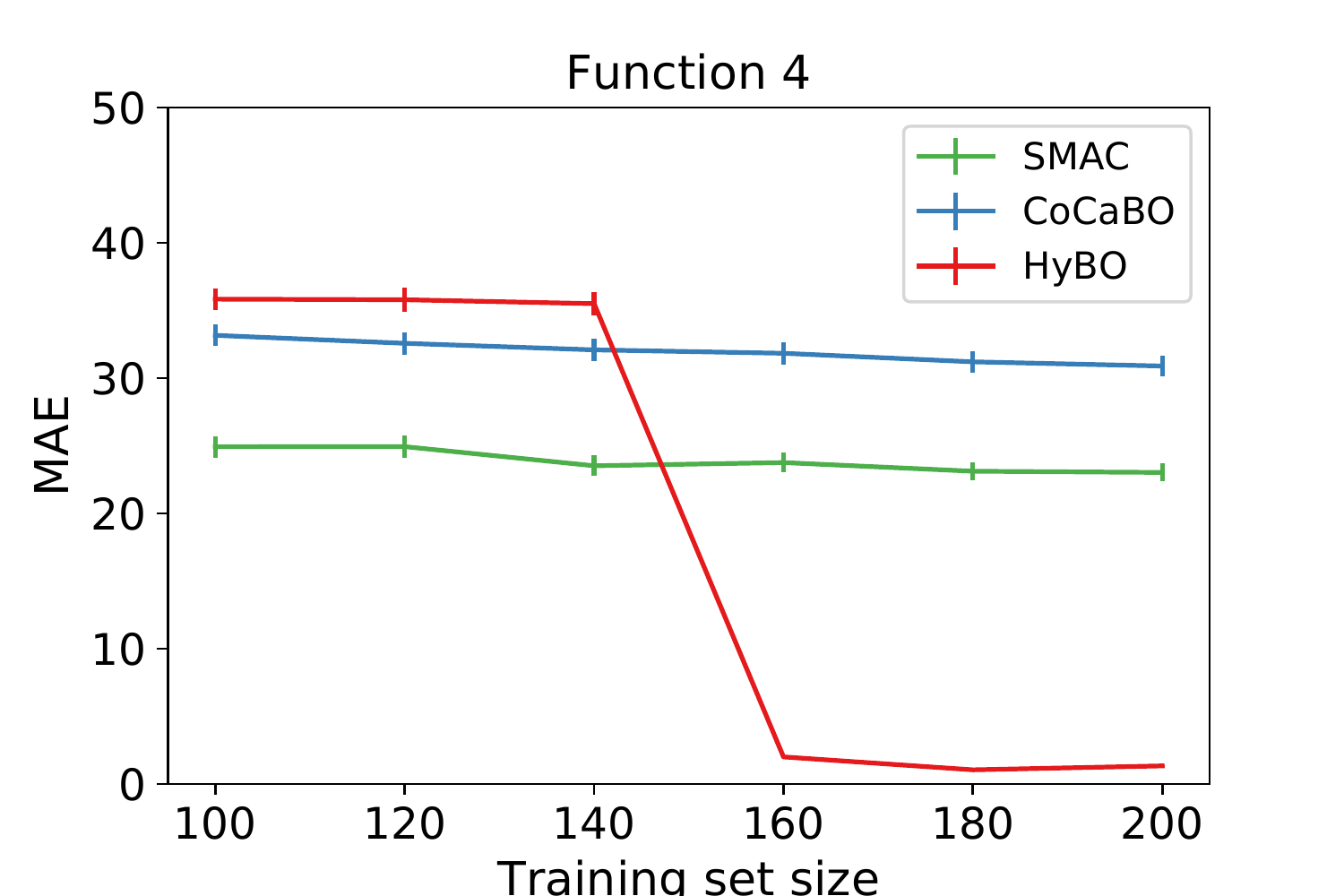}
\label{fig:mae_function_3}}
\caption{Results showing mean absolute test error with increasing size of training set on the bbob-mixint synthetic benchmarks.} 
\label{fig:mae}
\end{figure*}

\begin{figure*}[h!]
\centering
\subfloat[Subfigure 2 list of figures text][]{
\includegraphics[width=0.30\textwidth]{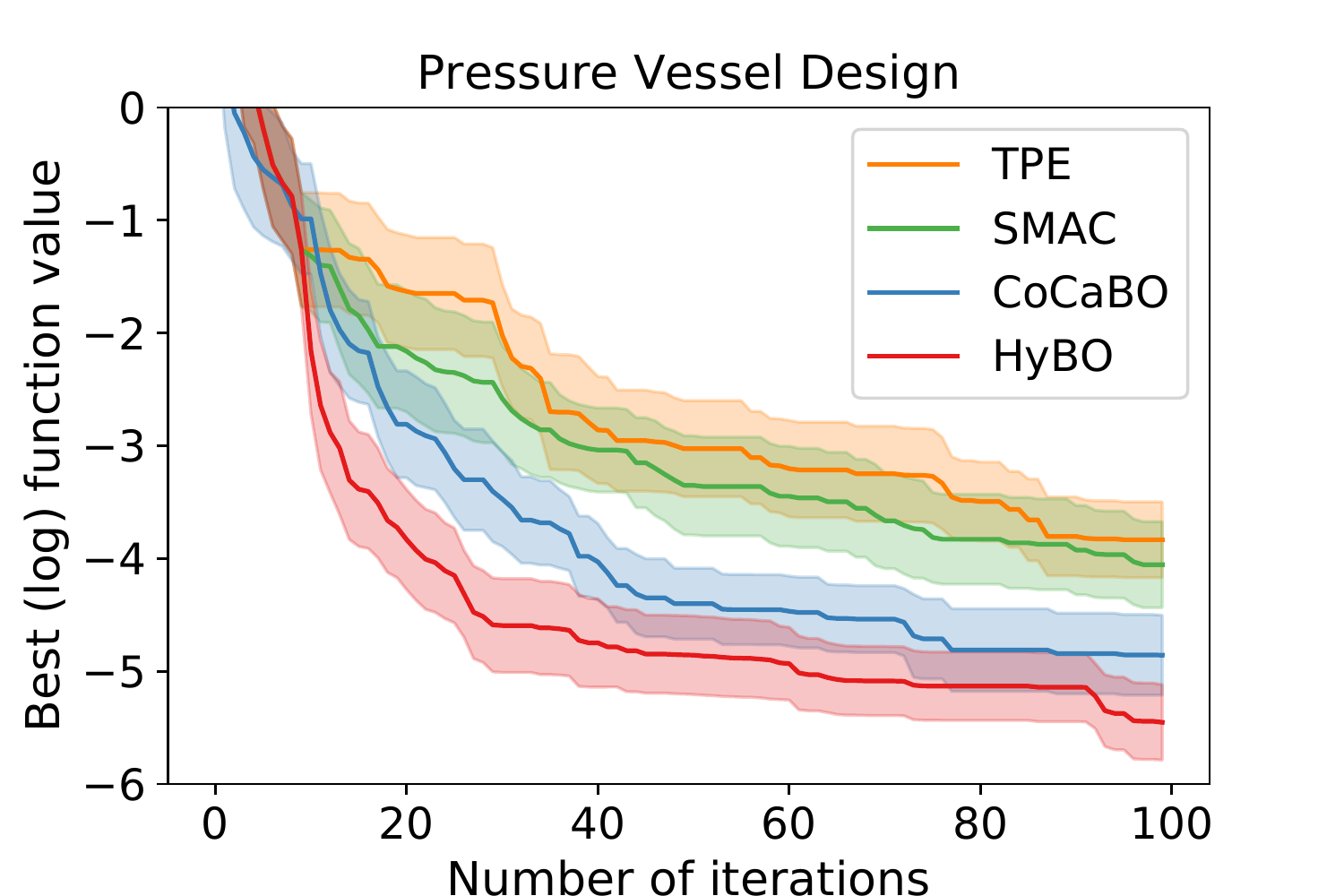}
\label{fig:pressure_vessel}}
\subfloat[Subfigure 2 list of figures text][]{
\includegraphics[width=0.30\textwidth]{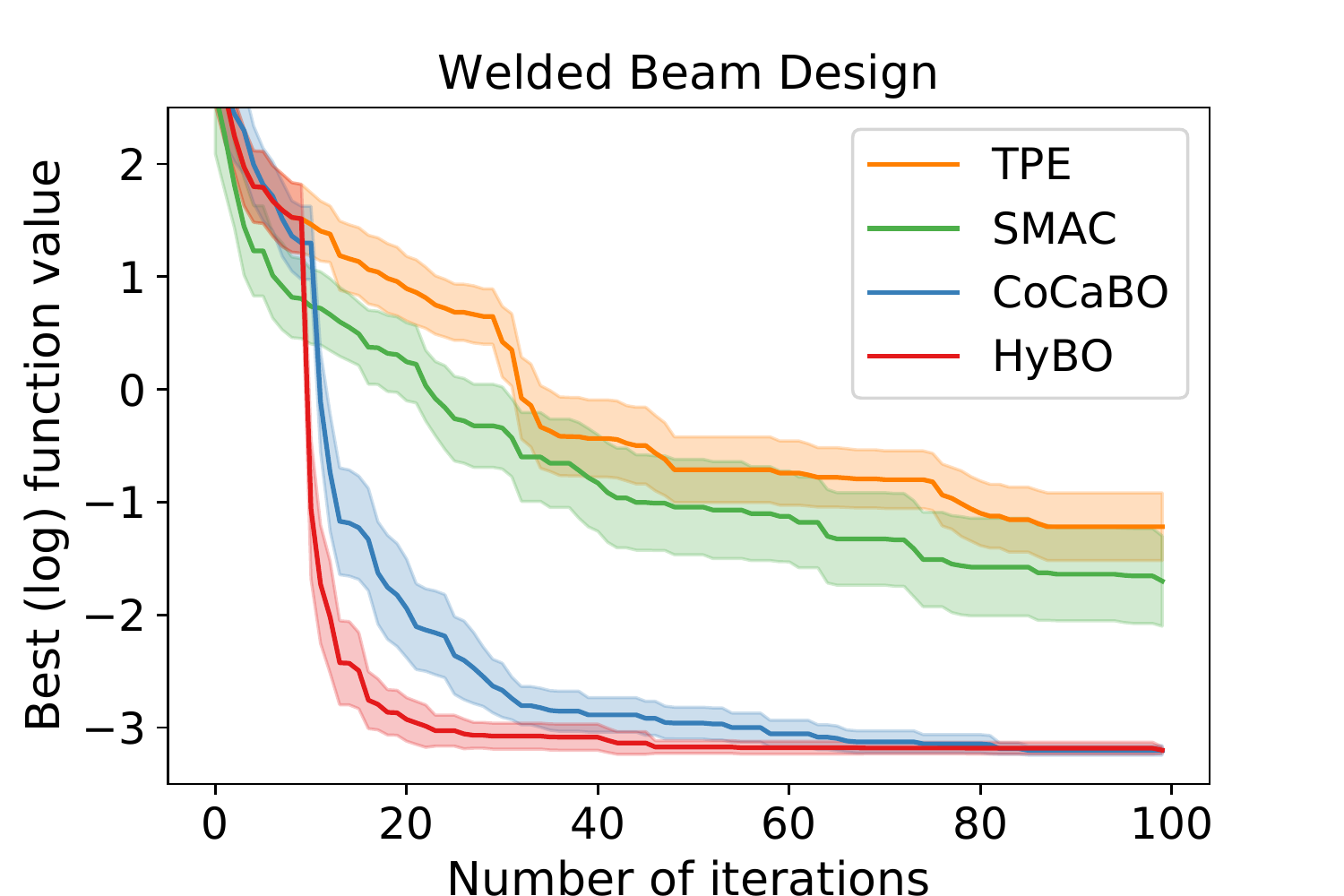}
\label{fig:weld_design}}
\subfloat[Subfigure 1 list of figures text][]{
\includegraphics[width=0.30\textwidth]{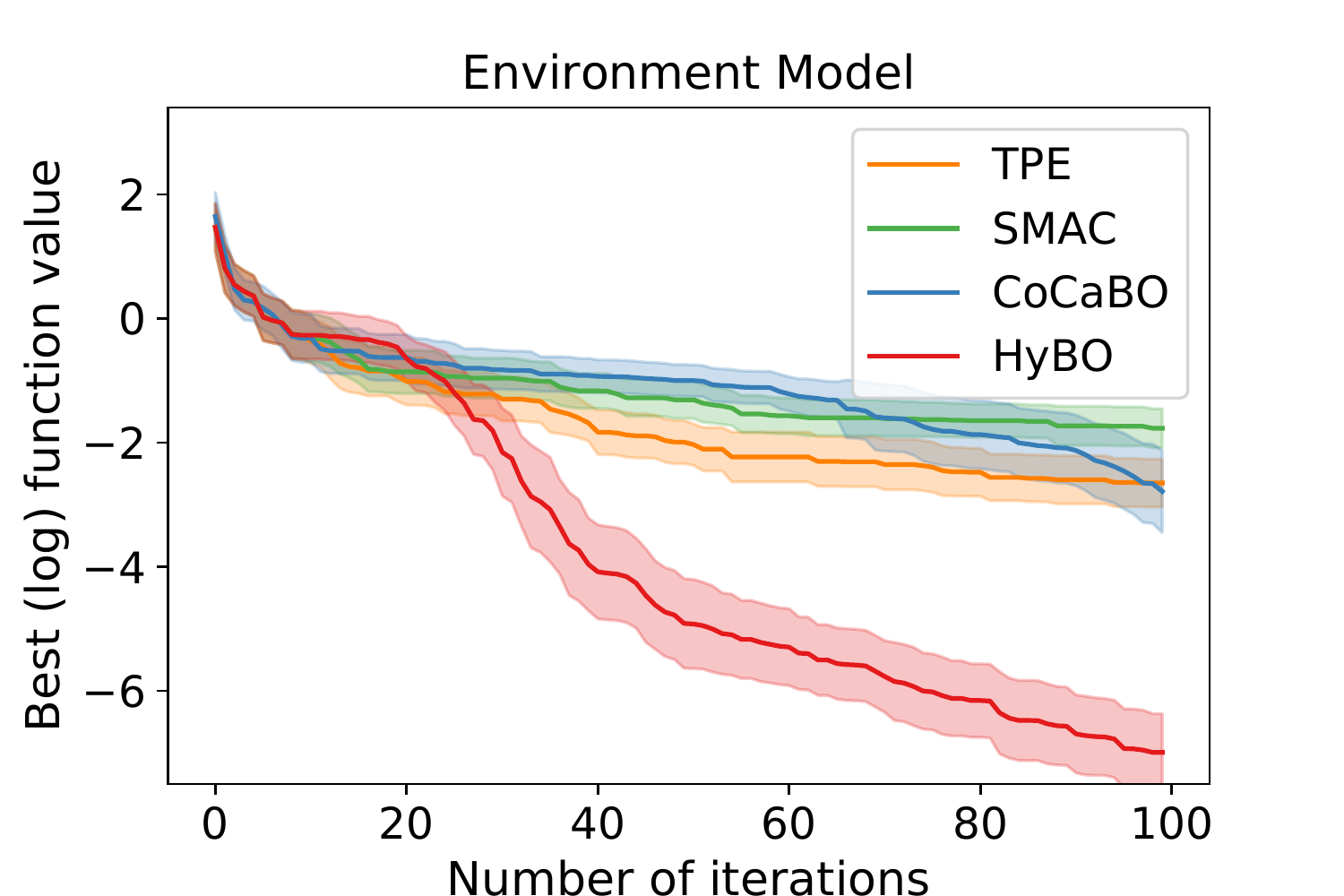}
\label{fig:em_func}}
\quad
\subfloat[Subfigure 1 list of figures text][]{
\includegraphics[width=0.30\textwidth]{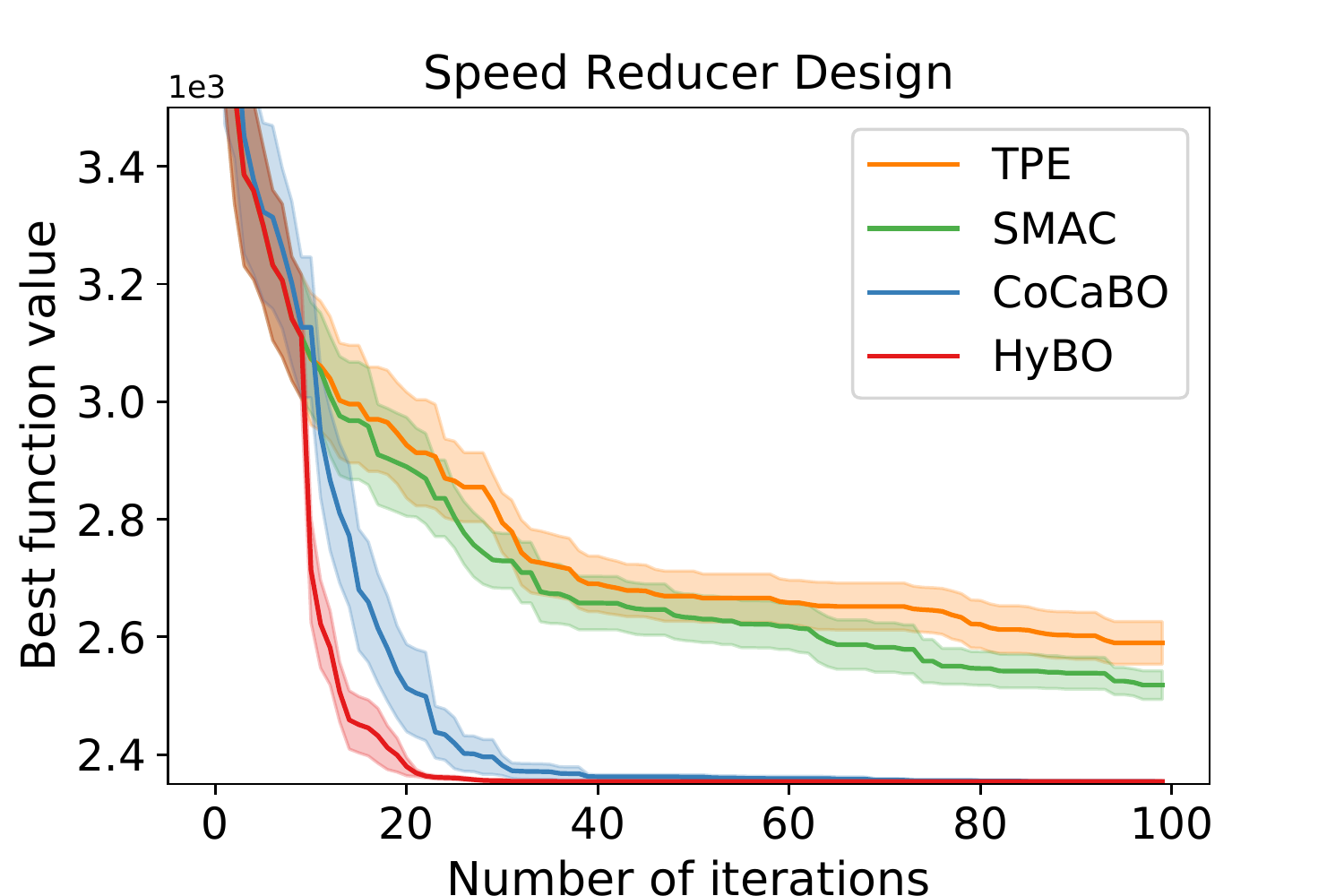}
\label{fig:speed_reducer}}
\subfloat[Subfigure 1 list of figures text][]{
\includegraphics[width=0.30\textwidth]{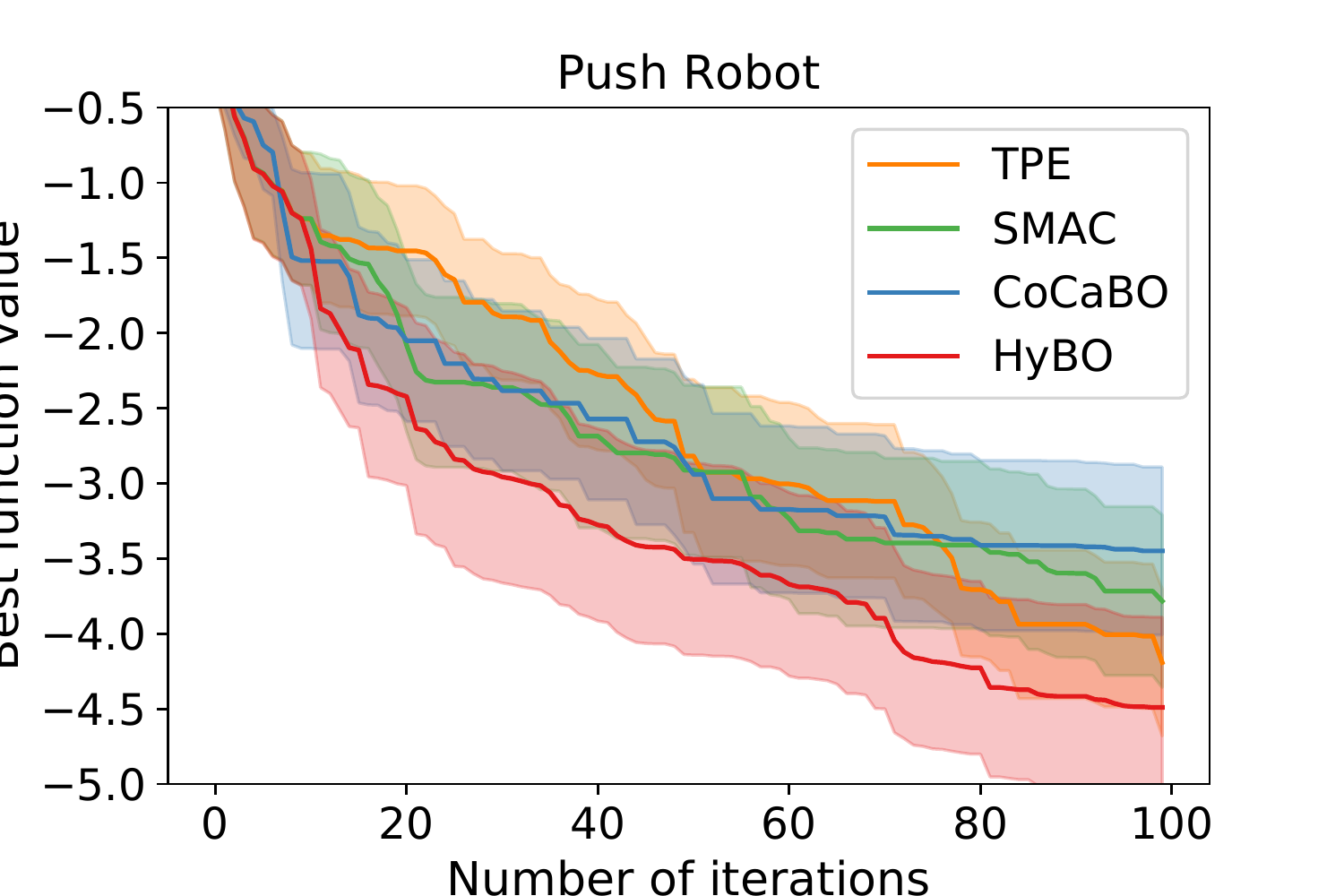}
\label{fig:push_robot}}
\caption{Results comparing the proposed HyBO approach with state-of-the-art baselines on multiple real world benchmarks.} 
\label{fig:real_world}
\end{figure*}

\begin{table*}[h!]
\centering
\begin{tabular}{|l | c | c | c | c | c |}  
\hline
{\bf Dataset} &  {\bf Cont-BO} & {\bf TPE} & {\bf SMAC}  & {\bf CoCaBO} & {\bf HyBO} \\
\hline
\hline
blood\_transfusion  & 76.089 (0.325) &  76.711 (0.432) & 76.658 (0.418) & 76.978 (0.455) &  {\bf 77.819 (0.463)}  \\
kc1  & 85.185 (0.129) &  85.637 (0.069) & 85.453 (0.087) & 85.415 (0.099) &  85.466 (0.116) \\
vehicle  & 80.501 (1.120) &  80.913  (1.051) & 83.669 (1.013) & 82.882  (1.222) & {\bf 86.104 (0.894)}   \\
segment  & 87.253 (0.995) &  87.792  (0.537) & 89.986  (0.692) & 89.639  (0.727) &  {\bf 91.433 (0.277)}  \\
cnae  & 95.370 (0.103) &  95.691  (0.082) & 95.605  (0.063) & 95.679  (0.108) &  95.644 (0.135)  \\
jasmine  & 77.317 (0.216) &  77.893  (0.071) & 77.460  (0.189) & 77.513  (0.202) &  77.121 (0.172)  \\
\hline
\end{tabular}
\caption{Results on the task of hyper-parameter tuning of neural network models. Bold numbers signify statistical significance.}
\label{tab:nn_hpo_results}
\end{table*}

\vspace{-1.5ex}

\noindent {\bf Results on  mixed integer benchmark suite.} Figure \ref{fig:synthetic} shows the canonical results on four benchmarks from \texttt{bbox-mixint} listed in Table \ref{tab:bbox} noting that all results show similar trends. HyBO and its variant HyBO-Round performs significantly better and converges much faster than all the other baselines. One key reason for this behavior is that hybrid diffusion kernel accounts for higher-order interactions between variables. Cont-BO performs the worst among all the methods. This shows that simply treating discrete variables as continuous is sub-optimal and emphasizes the importance of 
modeling the structure in discrete variables. 


\begin{table*}[h!]
\centering
\begin{tabular}{|l | c | c | c | c | c |}  
\hline
{\bf Benchmark} & {\bf TPE} & {\bf SMAC}  & {\bf CoCaBO} & {\bf HyBO} \\
\hline
Synthetic Function 1   & 0.012 & 2.34 & 2.30   & 50   \\
Synthetic Function 2   & 0.012 & 0.98 & 1.31   & 50   \\
Synthetic Function 3   & 0.026 & 2.99 & 3.18   & 180  \\
Synthetic Function 4   & 0.026 & 1.98 & 2.96   & 180  \\
Pressure Vessel Design & 0.003 & 0.34 & 0.85   & 20   \\
Welded Beam Design     & 0.004 & 0.64 & 1.02   & 40   \\
Speed Reducer Design   & 0.006 & 1.38 & 0.94   & 40   \\
Push Robot   & 0.017 & 1.94 &  1.70  &  90   \\
Environment model      & 0.005 & 0.31 & 0.50   & 40  \\
\hline
\end{tabular}
\caption{Computational cost in average wall-clock time (seconds) per BO iteration.}
\label{tab:computational_cost}
\end{table*}
\vspace{-1ex}
\noindent {\bf Ablation results for statistical models.} To understand the reasons for the better performance of HyBO, we compare the performance of its surrogate model based on hybrid diffusion kernels with those of CoCaBO and SMAC. We perform the following experiment. We constructed testing dataset (pairs of hybrid structures and their function evaluations) of size 200 via uniform random sampling. We compute the mean absolute error (MAE) of the three surrogate models as a function of training set size. The results are shown in Figure \ref{fig:mae} which depicts the mean and two times standard error of the MAE on 25 random testing datasets. HyBO clearly has very low error compared to CoCaBO and SMAC on Function 1 and 2. Although HyBO has similar MAE to CoCaBO in the beginning on Function 3 and 4, it rapidly decreases as the training set size increases which is not the case for other two methods. This experiment provides strong empirical evidence for the fact that the proposed surrogate model in HyBO can model hybrid spaces more accurately when compared to CoCaBO and SMAC.

\noindent {\bf Ablation results for marginalization in HyBO.} Bayesian treatment of hyper-parameters (marginalization) is one key component of our proposed HyBO method. However, to decouple the efficacy of additive diffusion kernel from the usage of marginalization, we performed experiments using HyBO without marginalization (HyBO w/o Marg in Figures). As evident from Figure \ref{fig:synthetic},  HyBO w/o Marg finds better solutions than all the baselines albeit with slower convergence which is improved by adding marginalization.

\noindent {\bf Results for real-world domains.} Figure \ref{fig:real_world} shows  comparison of HyBO approach with baseline methods on all real-world domains except hyper-parameter optimization. 
We make the following observations. 1) HyBO consistently performs better than all the baselines on all these benchmarks. 2) Even on benchmarks such as speed reducer design and welded beam design where HyBO finds a similar solution as CoCaBO, it does so with much faster convergence. 3) CoCaBO performs reasonably well on these benchmarks but its performance is worse than HyBO demonstrating that its sum kernel (along with Hamming kernel for discrete spaces) is less powerful than hybrid diffusion kernel of HyBO. 4). TPE has the worst performance on most benchmarks possibly a direct result of its drawback of not modeling the interactions between input dimensions. 5) SMAC performs poorly on all the benchmarks potentially due to poor uncertainty estimates from random forest surrogate model. 

Table \ref{tab:nn_hpo_results} shows the final accuracy (mean and standard error) obtained by all methods including HyBO on the task of tuning neural network models for six different datasets (BO curves are similar for all methods). HyBO produces comparable or better results than baseline methods.

\noindent {\bf Computational cost analysis.} We compare the runtime of different algorithms including HyBO. All experiments were run on a AMD EPYC 7451 24-Core machine. Table {\bf \ref{tab:computational_cost}} shows the average wall-clock time (in seconds) per BO iteration. We can see that HyBO is relatively expensive when compared to baseline methods. However, for real-world science and engineering applications, minimizing the cost of physical resources to perform evaluation (e.g., conducting an additive manufacturing experiment for designing materials such as alloys) is the most important metric. The computational cost for selecting inputs for evaluation is a secondary concern. HyBO uses more time to select inputs for evaluation to minimize the number of function evaluations to uncover better structures. We provide a finer-analysis of the HyBO runtime in Table \ref{tab:orders_of_interaction_time}. Each kernel evaluation time with all orders of interactions is very small. The overall runtime is spent on two major things: a) Sampling from posterior distributions of hyperparameters using slice sampling; and  b) AFO using CMA-ES + local search. We can reduce the sampling time by considering HyBO without marginalization which shows slightly worse performance, but takes only 10 percent of the sampling time in HyBO.

\begin{table}[h!]
\centering
\begin{tabular}{|l | c | c | c| c|}  
\hline
{\bf \specialcell{Orders of\\ interaction}} & {\bf \specialcell{ HyBO \\ iteration}} & {\bf \specialcell{AFO}}  & {\bf \specialcell{Sampling}} & {\bf \specialcell{Kernel \\ eval.}} \\
\hline
2 & 62 & 46 & 16 & 0.005   \\
5 &  68 & 50 & 18 & 0.006   \\
10  &  102 & 68 & 34 & 0.010     \\
20 (HyBO) &  180 & 114 & 66 & 0.020   \\
\hline
\end{tabular}
\vspace{-2ex}
\caption{Average runtime (seconds) for different orders of interaction within hybrid kernel for synthetic Function 3.}
\label{tab:orders_of_interaction_time}
\end{table}

\vspace{-3.0ex}

\section{Conclusions}
We studied a novel Bayesian optimization approach referred as HyBO for optimizing hybrid spaces using Gaussian process based surrogate models. We presented a principled approach to construct hybrid diffusion kernels by combining diffusion kernels defined over continuous and discrete sub-spaces in a tractable and flexible manner to capture the interactions between discrete and continuous variables. We proved that additive hybrid kernels have the universal approximation property. 
Our experimental results on diverse synthetic and real-world benchmarks show that HyBO performs significantly better than state-of-the-art methods.

\noindent {\bf Acknowledgements.} This research is supported 
by NSF grants IIS-1845922, OAC-1910213, and CNS-1955353. 

\appendix
\clearpage
\section{Appendix}
In this section, we illustrate the additive hybrid diffusion kernel (Equation \ref{eqn:additive_diff_kernel}) by providing a running example. 

\subsection{Running example for additive hybrid diffusion kernel}
We illustrate the additive hybrid diffusion kernel and its recursive computation using a 3-dimensional hybrid space, where the first two dimensions correspond to discrete subspace and the last dimension correspond to continuous subspace. Let $k_1, k_2, k_3$ be the base kernels for first, second, and third dimension respectively. The additive diffusion kernel can be computed recursively step-wise as shown below:
\begin{align*}
\mathcal{K}_1 &=  \theta_1^2 \cdot (k_1 + k_2 + k_3), \hspace{13mm} \textcolor{red}{\mathcal{S}_1 =   (k_1 + k_2 + k_3)}\\
\mathcal{K}_2 &=  \theta_2^2 \cdot (k_1 k_2 + k_1 k_3 + k_2 k_3), \hspace{2.5mm} \textcolor{red}{\mathcal{S}_2 =(k_1^2 + k_2^2 +  k_3^2)}\\
\mathcal{K}_3 &=  \theta_3^2 \cdot (k_1 k_2 k_3), \hspace{21mm} \textcolor{red}{\mathcal{S}_3 =   (k_1^3 + k_2^3 + k_3^3)}\\ 
\mathcal{K}_0 &= 1; \\
\mathcal{K}_1 &= \theta_1^2 \cdot \textcolor{red}{S_1}; \\
\mathcal{K}_2 &= \theta_2^2 \cdot \frac{1}{2}\left(\mathcal{K}_1 \cdot \textcolor{red}{S_1} - \textcolor{red}{S_2}\right); \\
\mathcal{K}_3 &= \theta_3^2 \cdot \frac{1}{3}\left(\mathcal{K}_2 \cdot \textcolor{red}{S_1} - \mathcal{K}_1 \cdot \textcolor{red}{S_2} + \textcolor{red}{S_3}\right); \\
\mathcal{K}_{HYB} &= \mathcal{K}_1  + \mathcal{K}_2 + \mathcal{K}_3
\end{align*}

\section{Additional Experimental Details}
\subsection{Real world benchmarks}
{\bf 1) Pressure vessel design optimization.} The objective function (cost of cylindrical pressure vessel design) $\mathcal{F}(x)$ for this domain is given below:
\begin{align}
\begin{split}
    \min_{\{x_1, x_2, x_3, x_4\}}& 0.6224x_1x_3x_4 + 1.7781x_2x_3^2  + \\
    & 3.1661 x_1^2x_4 + 19.84 x_1^2 x_3
\end{split}
\end{align}
where $x_1, x_2$ are discrete variables (thickness of shell and head of pressure vessel) lying in $\{1, \cdots, 100\}$ and $x_3 \in [10, 200], x_4 \in [10, 240]$ are continuous variables (inner radius and length of cylindrical section). 


{\bf  2) Welded beam design optimization.}   The objective function (cost of fabricating welded beam) $\mathcal{F}(x)$ for this domain is: 

\begin{align}
     \min_{\{x_1, x_2, x_3, x_4, x_5, x_6\}} (1+G_1) (x_1 x_5 + x_4) x_3^2 + G_2 x_5 x_6 (L + x_4)
\end{align} 
where $x_1 \in \{0, 1\}, x_2 \in  \{0, 1, 2, 3\}$ are discrete variables, $x_3 \in [0.0625, 2], x_4 \in [0, 20], x_5 \in [2, 20], x_6 \in [0.0625, 2]$ are continuous variables, $G_1$ is the cost per volume of the welded material, and $G_2$ is the cost per volume of the bar stock. The constants ($G_1, G_2, L$), which are dependent on the second discrete variable $x_2$, are given in \cite{weld_design_1,weld_design_2}.


{\bf 3) Speed reducer design optimization.} The objective function (weight of speed reducer) $\mathcal{F}(x)$ for this domain is:
\begin{align}
\begin{split}
    &\min_{\{x_1, x_2, x_3, x_4, x_5, x_6, x_7\}}  0.79 x_2 x_3^2(3.33 x_1^3 + 14.93 x_1 - 43.09) \\
    & -1.51 x_2 (x_6^2 + x_7^2) + 7.48 (x_6^3 + x_7^3) + 0.79 (x_4 x_6^2 + x_5 x_7^2)
\end{split}
\end{align}
where $x_1 \in \{17,18 \cdots,28\}$ represents the discrete variable (number of teeth on pinion), $x_2 \in [2.6, 3.6], x_3  \in [0.7, 0.8], x_4 \in [7.3, 8.3], x_5 \in [0.7, 0.8], x_6 \in [2.9, 3.9], x_7 \in [5, 5.5]$ represents the continuous variables  (face width, teeth module, lengths of shafts between bearings, and diameters of the shafts respectively).


The above three benchmarks are usually described with {\em known} constraints in a declarative manner. However, for simplicity, we consider their unconstrained version for evaluation in this paper. If required, since the constraints are known, we can easily avoid searching for invalid solutions by using an appropriate acquisition function optimizer within HyBO.

{\bf 4) Optimizing control for robot pushing.} This domain was taken from this URL \footnote{\url{https://github.com/zi-w/Ensemble-Bayesian-Optimization/tree/master/test_functions}}. We consider a hybrid version of this problem by discretizing the location parameters ($x_1, x_2, x_3, x_4 \in \{-5, -4, \cdots, 5\}$ and $x_5, x_6, x_7, x_8 \in \{-10, -9, \cdots, 10\}$). There are two other discrete variables corresponding to simulation steps $x_9, x_{10} \in \{2, 3, 4, \cdots, 30\}$ and two continuous variables $x_{11}, x_{12}$ lying in $[0, 2\pi]$.


{\bf 5) Calibration of environmental model.} The details of the objective function for this domain are available in \cite{em_func, astudillo2019bayesian}. The single  discrete variable has 284 candidate values lying in the set $\{30.01, 30.02, \cdots 30.285\}$. There are three continuous variables lying in the range: $x_2 \in [7,13], x_3 \in [0.02, 0.12], x_4 \in [0.01, 3]$. 


{\bf 6) Hyper-parameter optimization.} The type and range for different hyper-parameters considered in this domain are given in Table \ref{tab:hyper_ranges}. We employed the scikit-learn \cite{scikit-learn} neural network implementation for this benchmark.

\begin{table*}[t!]
    \centering
    \begin{tabular}{|c|c|c|}
    \hline
    Hyperparameter & Type & Range \\
    \hline \hline
     Hidden layer size    &  Discrete  & $\{40, 60, \cdots, 300\}$\\
    Type of activation & Discrete & $\{\text{'identity'}, \text{'logistic'}, \text{'tanh'}, 
     \text{'relu'}\}$\\
     Batch size & Discrete & $\{40, 60, \cdots, 200\}$ \\
     Type of learning rate & Discrete & $\{\text{'constant'}, \text{'invscaling'}, \text{'adaptive'}\}$ \\
     Early stopping & Discrete & True/False \\
     Learning rate initialization & Continuous & $[0.001, 1]$ \\
     Momentum & Continuous & $[0.5, 1]$ \\
     Alpha parameter & Continuous & $[0.0001, 1]$ \\
     \hline
    \end{tabular}
    \caption{Type and range of hyper-parameters considered for the HPO benchmark.}
    \label{tab:hyper_ranges}
\end{table*}

\begin{figure*}[h!]
\centering
\subfloat[Subfigure 2 list of figures text][]{
\includegraphics[width=0.30\textwidth]{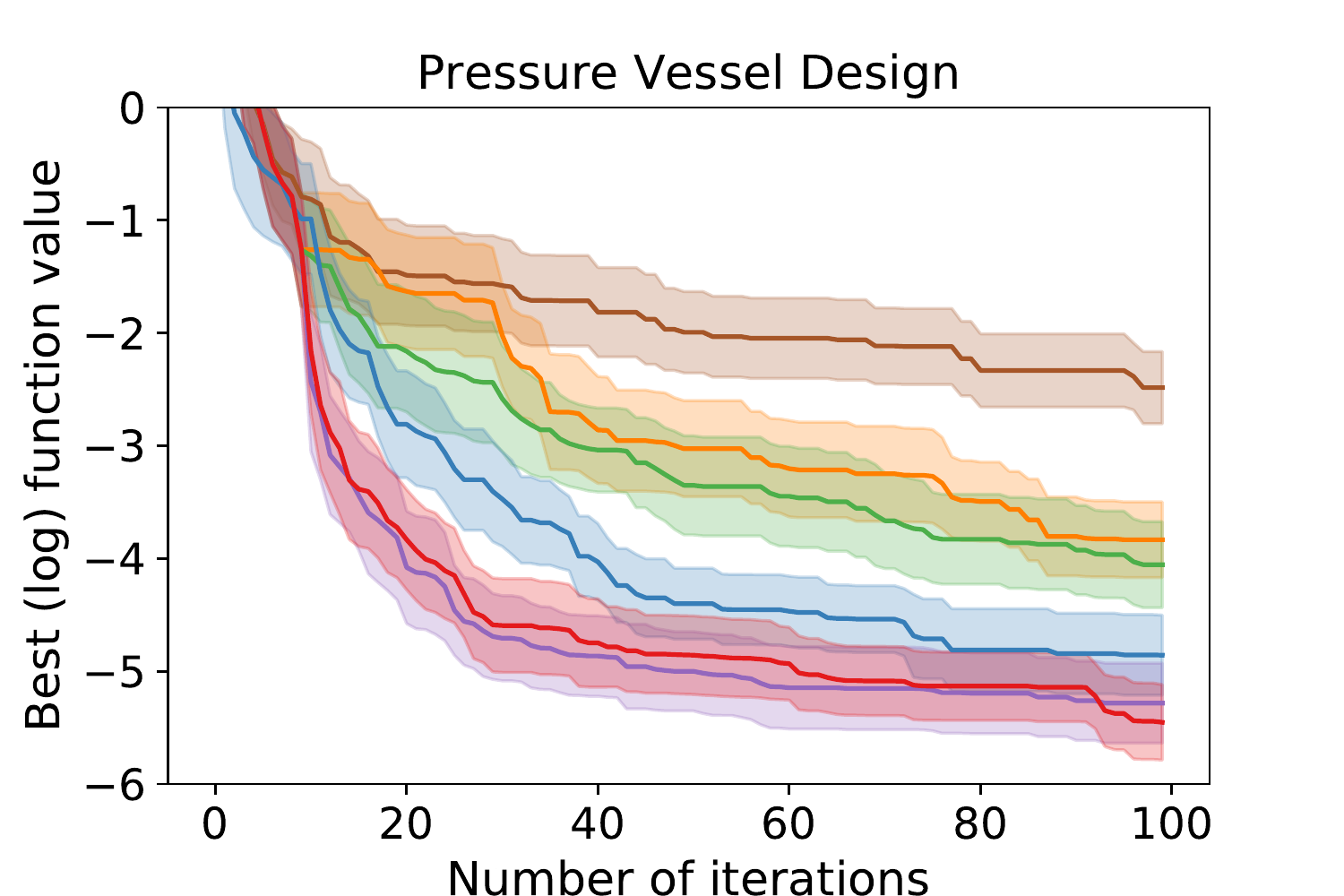}
\label{fig:pressure_vessel}}
\subfloat[Subfigure 2 list of figures text][]{
\includegraphics[width=0.30\textwidth]{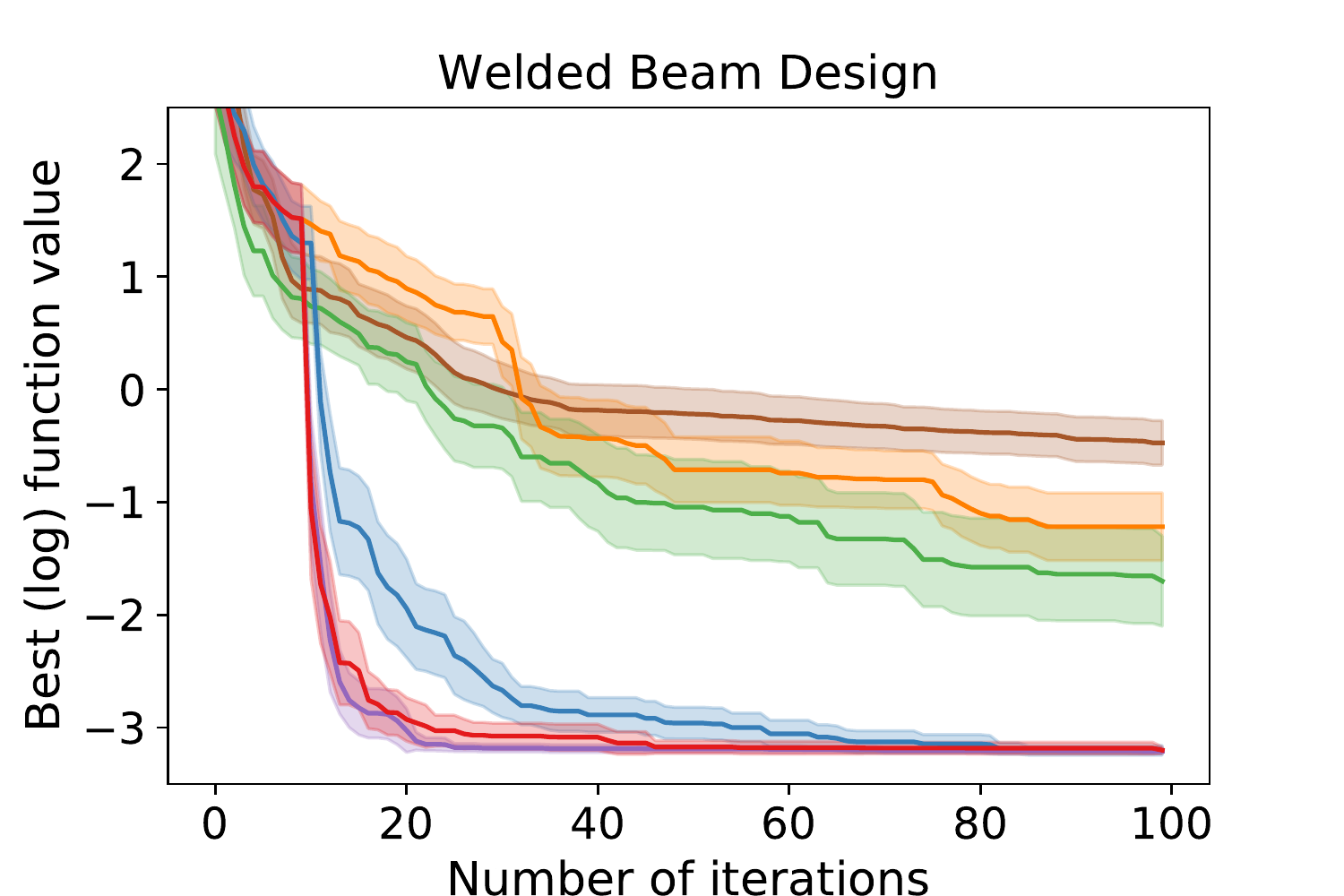}
\label{fig:weld_design}}
\subfloat[Subfigure 1 list of figures text][]{
\includegraphics[width=0.30\textwidth]{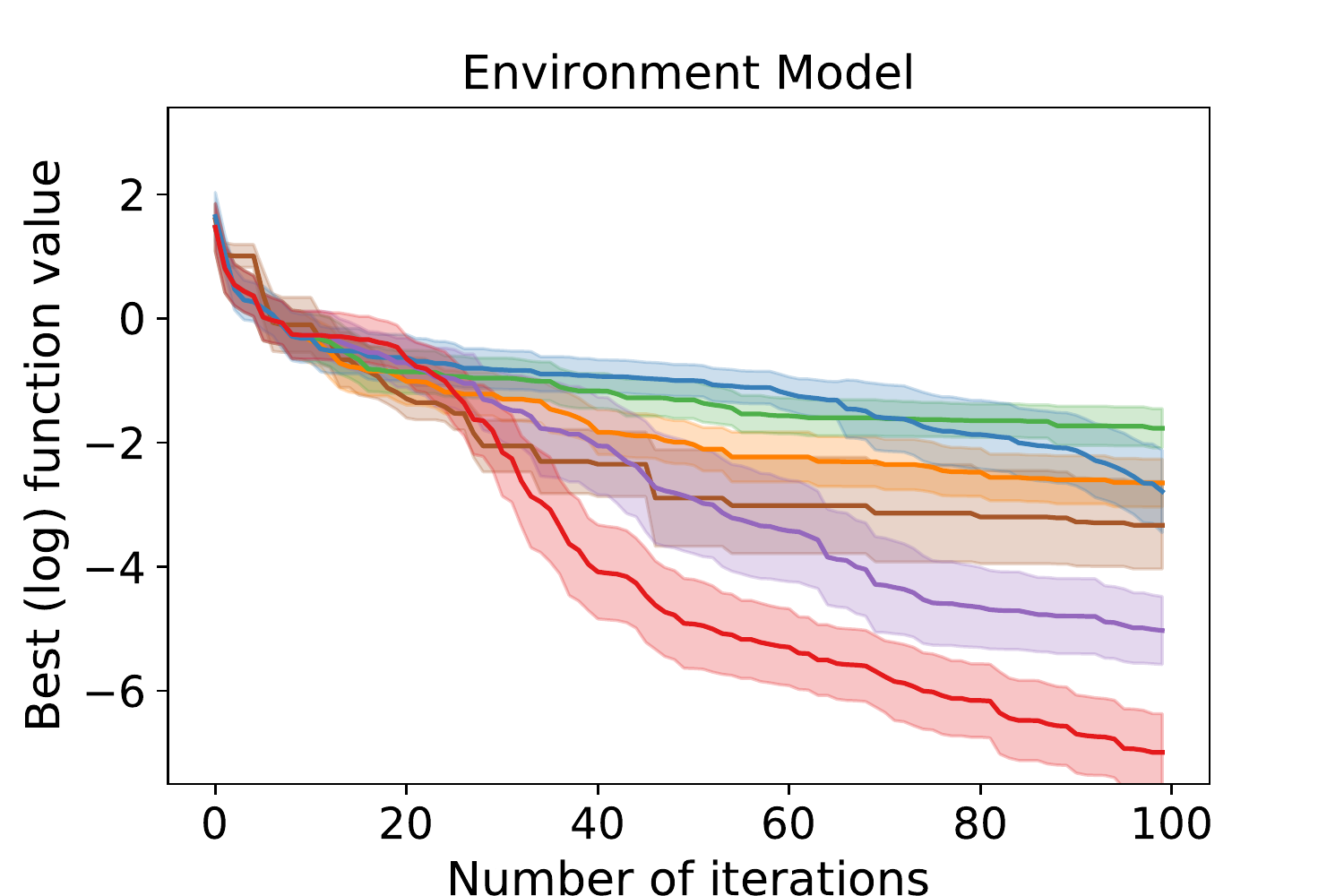}
\label{fig:em_func}}
\quad
\subfloat[Subfigure 1 list of figures text][]{
\includegraphics[width=0.30\textwidth]{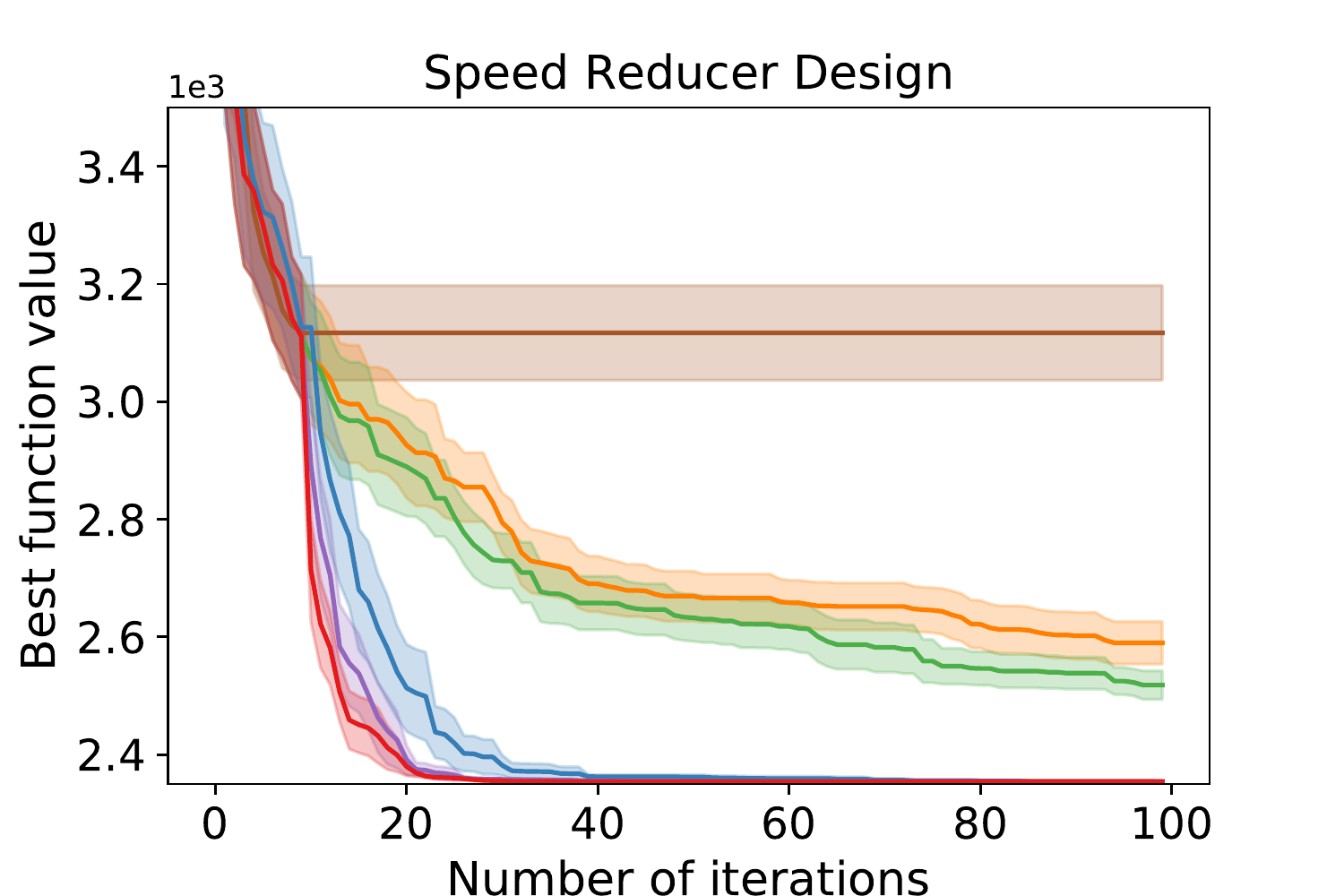}
\label{fig:speed_reducer}}
\subfloat[Subfigure 1 list of figures text][]{
\includegraphics[width=0.30\textwidth]{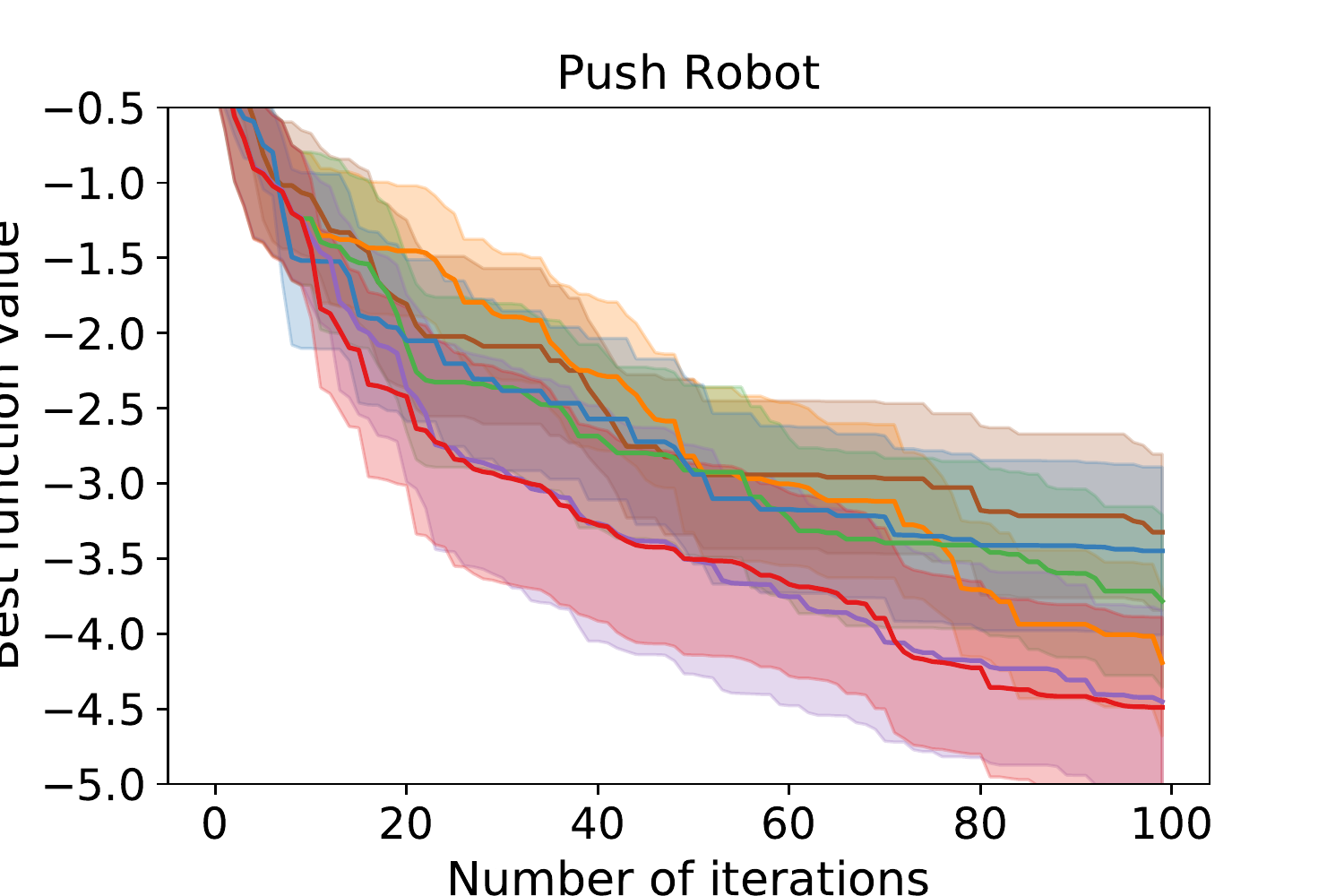}
\label{fig:hybo_legend}}
\includegraphics[width=0.30\textwidth]{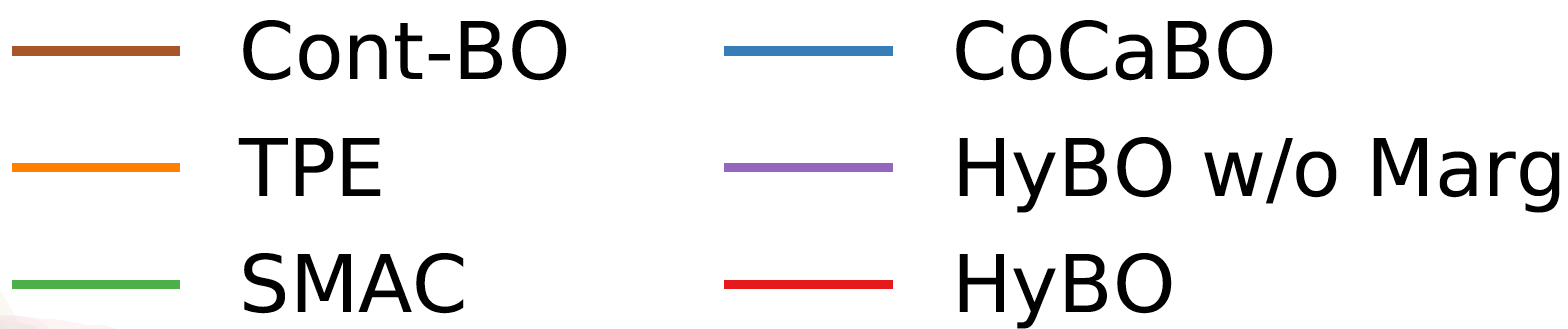}
\label{fig:hybo_legend}
\caption{Results comparing the proposed HyBO approach with state-of-the-art baselines on multiple real world benchmarks. These figures also contain HyBO without marginalization and Cont-BO results.} 
\label{fig:real_world_appendix}
\end{figure*}

\section{Additional Results}
\noindent {\bf Results for real-world benchmarks.} Figure \ref{fig:real_world_appendix} extends the plots of Figure \ref{fig:real_world} by including the performance of Cont-BO and HyBO w/o Marg on the real-world benchmarks. The results show similar trend where Cont-BO performs worse than all other methods showing the need to take into account the hybrid input structure. Also, the performance of HyBO w/o Marg remains similar to HyBO (except on calibration of environment model) demonstrating the effective modeling strength of additive hybrid diffusion kernel.

\begin{table}[h!]
\centering
\begin{tabular}{|l | c | c | c|}  
\hline
{\bf Benchmark} & {\bf HyBO} & {\bf G-M et al.}  & {\bf Vanilla BO} \\
\hline
Synthetic Function 1   & 79.7 & 99.4 & 86.2   \\
Synthetic Function 2   & 394.6 & 420 & 407   \\
Synthetic Function 3   & 81.1 & 143 & 135    \\
Synthetic Function 4   & 395.2 & 458 & 456.8  \\
\hline
\end{tabular}
\caption{Results for additional baseline experiments}
\label{tab:more_baselines}
\end{table}

\noindent{\bf Comparison with \cite{lobato}}  
As mentioned in our related work, this is an interesting approach for BO over discrete spaces but it is specific to discrete spaces alone. Since our problem setting considers hybrid input spaces, we performed experiments using this method for the discrete part and using the standard BO approach for the continuous part with HyBO’s AFO procedure.  Results of this approach (referred as G-M et al.,) on the 4 synthetic benchmarks are shown in Table \ref{tab:more_baselines}. The best function value achieved after 200 iterations and averaged over 25 different runs (same configuration as described in the main paper) is shown. We also add another baseline named Vanilla BO (GP with RBF kernel to model hybrid space + HyBO’s AFO procedure) in Table \ref{tab:more_baselines}. It is evident from the results that HyBO performs significantly better.

\clearpage 


\bibliography{main}
\bibliographystyle{icml2021}






\end{document}